\documentclass[11pt, a4paper, singlecolumn, confidential, copyright, gdm]{google}
\renewcommand{\internalonly}{}

\usepackage{graphicx}
\usepackage{booktabs}
\usepackage{makecell}   
\usepackage{array}
\usepackage{subcaption}
\usepackage{eccvabbrv}

\usepackage{standalone} 
\usepackage{tikz}
\usetikzlibrary{arrows.meta, decorations.markings, calc, backgrounds, positioning}
\usepackage{amsmath} 
\usepackage{bm}

\usepackage{pifont} 
\usepackage{wrapfig}
\usetikzlibrary{
  arrows.meta, positioning, calc, backgrounds, fit, decorations.pathreplacing
}

\definecolor{cU}{RGB}{220,60,45}
\definecolor{cEmbed}{RGB}{52,152,219}
\definecolor{cHead}{RGB}{142,68,173}
\definecolor{cStudent}{RGB}{243,156,18}
\definecolor{cGreen}{RGB}{39,174,96}
\definecolor{cGray}{RGB}{100,110,120}
\usepackage[numbers]{natbib}

\usepackage{orcidlink}

\usepackage{physics}
\usepackage{multirow}

\usepackage{microtype} 
\usepackage{cleveref}  

\usepackage{algorithm}
\usepackage{algpseudocode}
\usepackage[utf8]{inputenc}

\usepackage{listings}
\usepackage{xcolor}

\definecolor{codegreen}{rgb}{0.2,0.6,0.4}
\definecolor{codeblue}{rgb}{0.1,0.3,0.8}
\definecolor{codemagenta}{rgb}{0.8,0.0,0.5}

\lstdefinestyle{jaxstyle}{
    language=Python,
    backgroundcolor=\color{white},   
    commentstyle=\color{codegreen},
    keywordstyle=\color{codeblue},
    stringstyle=\color{codemagenta},
    basicstyle=\ttfamily\small,
    breakatwhitespace=false,         
    breaklines=true,                 
    captionpos=t,                    
    keepspaces=true,                 
    numbers=none,                    
    showspaces=false,                
    showstringspaces=false,
    showtabs=false,                  
    tabsize=4,
    columns=fullflexible,
    morekeywords={jax, lax, cond, fori_loop, stop_gradient, randint, task_loss, dist_loss},
    literate={+}{{\textcolor{codeblue}{+}}}1
             {-}{{\textcolor{codeblue}{-}}}1
             {*}{{\textcolor{codeblue}{*}}}1
             {=}{{\textcolor{codeblue}{=}}}1
             {/}{{\textcolor{codeblue}{/}}}1
             {<}{{\textcolor{codeblue}{<}}}1,
}
\newcommand{\ours}{ELT}
\newcommand{\oursfull}{Elastic Looped Transformers}

\newcommand{\distill}{ILSD}
\newcommand{\distillfull}{Intra-Loop Self Distillation (ILSD)}


\uselogo{} 

\title{\ours: \oursfull\ for Visual Generation} 

\reportnumber{} 

\renewcommand{\today}{}

\author[*]{Sahil Goyal}
\author[*]{Swayam Agrawal}
\author[ ]{Gautham Govind Anil}
\author[ ]{Prateek Jain}
\author[$\dagger$]{Sujoy Paul}
\author[$\dagger$]{Aditya Kusupati}
\affil[*]{Equal contributions}
\affil[$\dagger$]{Equal advising}

\teaser{
    \begin{figure}[ht]
    \centering
    \begin{minipage}[b]{0.48\linewidth}
\captionsetup{justification=centering}
        \centering
        \resizebox{\textwidth}{!}
    {
    \setlength{\tabcolsep}{1pt}
    
    \begin{tabular}{ccccc}
    $L = 2$ & $L = 4$ & $L = 6$ & $L = 8$ & $L = 10$ \\
      \includegraphics[width=0.195\linewidth]{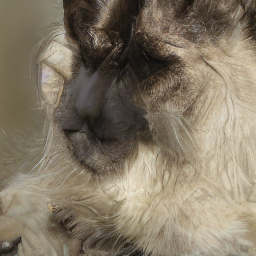} & \includegraphics[width=0.195\linewidth]{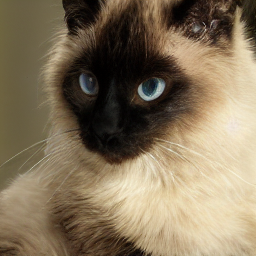} &
        \includegraphics[width=0.195\linewidth]{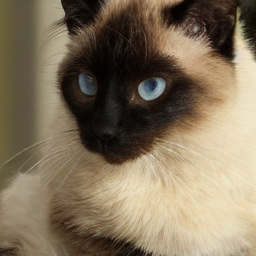} &
        \includegraphics[width=0.195\linewidth]{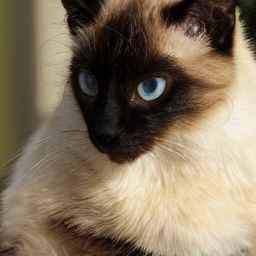} &
        \includegraphics[width=0.195\linewidth]{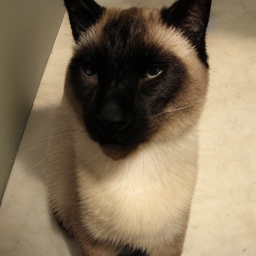}\\[-2pt]
        \includegraphics[width=0.195\linewidth]{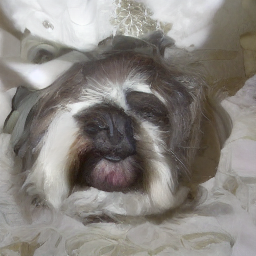} & \includegraphics[width=0.195\linewidth]{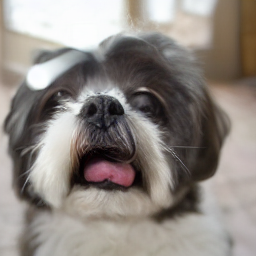} &
        \includegraphics[width=0.195\linewidth]{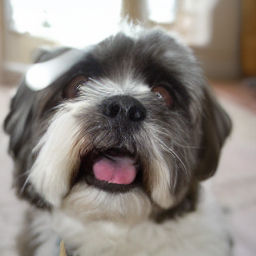} &
        \includegraphics[width=0.195\linewidth]{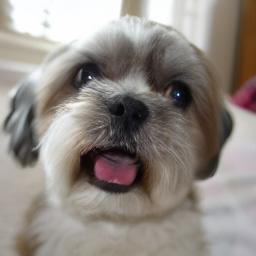} &
        \includegraphics[width=0.195\linewidth]{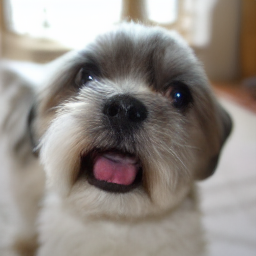}\\[-2pt]
        \includegraphics[width=0.195\linewidth]{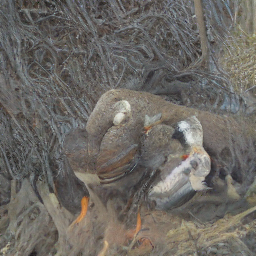} & \includegraphics[width=0.195\linewidth]{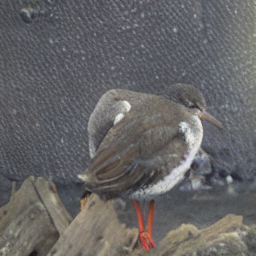} &
        \includegraphics[width=0.195\linewidth]{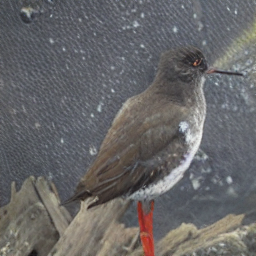} &
        \includegraphics[width=0.195\linewidth]{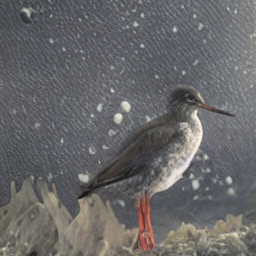} &
        \includegraphics[width=0.195\linewidth]{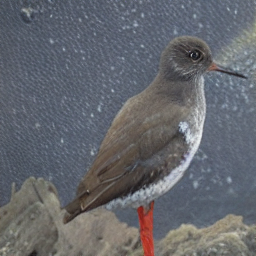}\\
    \end{tabular}
    }
        \caption*{ \footnotesize (a) \textbf{\ours}'s \textbf{Any-Time inference} from \textbf{single training run}}
    \end{minipage}%
    \hfill
    \begin{minipage}[b]{0.48\linewidth}
    \captionsetup{justification=centering}
        \centering
        \resizebox{\textwidth}{!}
    {
    \setlength{\tabcolsep}{1pt}
    
    \begin{tabular}{ccccc}
    $L = 2$ & $L = 4$ & $L = 6$ & $L = 8$ & $L = 10$ \\
        \includegraphics[width=0.195\linewidth]{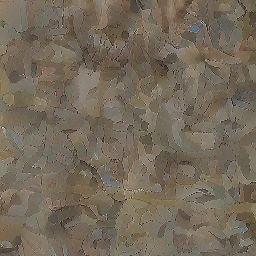} &  \includegraphics[width=0.195\linewidth]{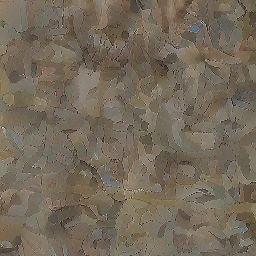} &
        \includegraphics[width=0.195\linewidth]{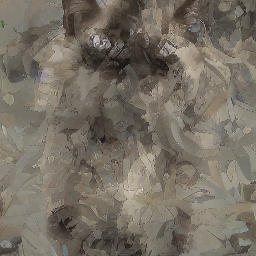} &
        \includegraphics[width=0.195\linewidth]{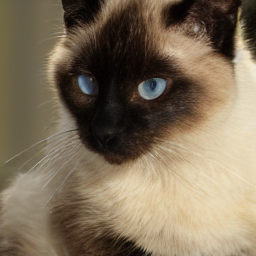} &
        \includegraphics[width=0.195\linewidth]{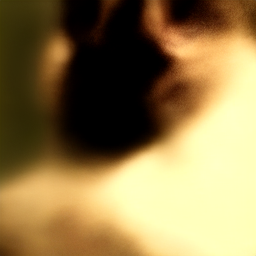}\\[-2pt]
        \includegraphics[width=0.195\linewidth]{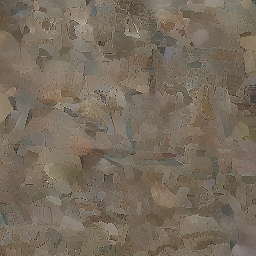} & \includegraphics[width=0.195\linewidth]{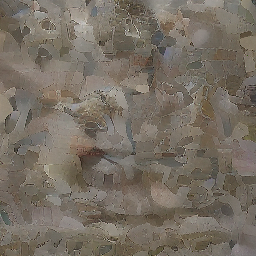} &
        \includegraphics[width=0.195\linewidth]{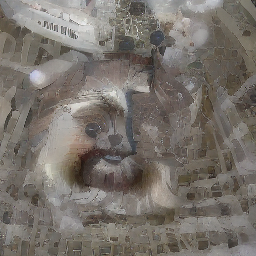} &
        \includegraphics[width=0.195\linewidth]{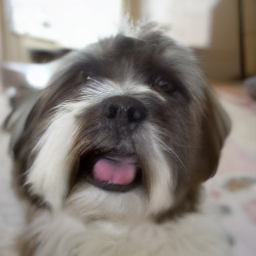} &
        \includegraphics[width=0.195\linewidth]{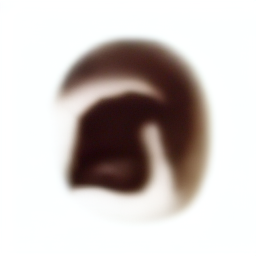}\\[-2pt]
        \includegraphics[width=0.195\linewidth]{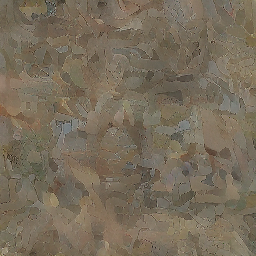} & \includegraphics[width=0.195\linewidth]{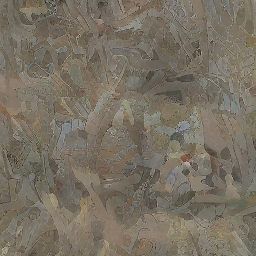} &
        \includegraphics[width=0.195\linewidth]{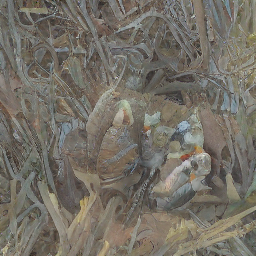} &
        \includegraphics[width=0.195\linewidth]{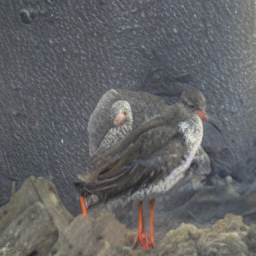} &
        \includegraphics[width=0.195\linewidth]{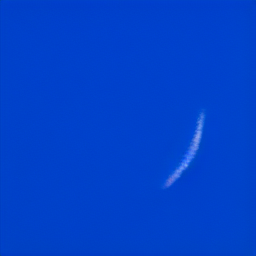}\\
    \end{tabular}
    }
        \caption*{\footnotesize (b) Vanilla looped transformers}
    \end{minipage}%

    \caption{\small \textbf{Class-conditional Image Generation on ImageNet $\bm{256 \times 256}$}. Comparing our proposed \oursfull\ (left) against vanilla looped transformers (right), which simply repeat transformer layers during a forward pass of diffusion. As shown, standard looping only produces a coherent image when the inference loop count exactly matches its training setup ($L=8$), with quality severely degrading at all other values. In contrast, our approach maintains high-fidelity generation across several evaluated compute budgets.
    }
    
    \label{fig:gen_dit_ilsd}
    \vspace{-1cm}
    
\end{figure}
}

\begin{abstract}

We introduce \oursfull\ (\ours), a highly parameter-efficient class of visual generative models based on a recurrent transformer architecture. While conventional generative models rely on deep stacks of unique transformer layers, our approach employs iterative, weight-shared transformer blocks to drastically reduce parameter counts while maintaining high synthesis quality. To effectively train these models for image and video generation, we propose the idea of \distillfull, where student configurations (intermediate loops) are distilled from the teacher configuration (maximum training loops) to ensure consistency across the model's depth in a single training step. 
Our framework yields a family of elastic models from a single training run, enabling Any-Time inference capability with dynamic trade-offs between computational cost and generation quality, with the same parameter count. \ours\ significantly shifts the efficiency frontier for visual synthesis. With \emph{4}$\times$ reduction in parameter count under iso-inference-compute settings, \ours\ achieves a competitive FID of \emph{2.0} on class-conditional ImageNet $256 \times 256$ and FVD of  \emph{72.8} on class-conditional UCF-101.
\end{abstract}

\begin{document}
\maketitle

\section{Introduction}
\label{sec:intro}

Traditional techniques to increase compute capacity in deep learning models, such as stacking deeper layers or increasing network width, inevitably lead to a proportionally larger memory footprint. Recurrence offers a powerful alternative paradigm, enabling models to utilize large amounts of compute while maintaining a minimal memory footprint by leveraging the same set of parameters repeatedly. 
This architectural efficiency parallels the biological visual systems~\citep{kar2019evidence,kietzmann2019recurrence}, where recurrent processing, rather than strictly feedforward pathways, is essential for resolving complex visual inputs.

While looping of transformers was popularized by Universal Transformers~\citep{dehghani2018universal} and has recently empowered language models with stronger reasoning capabilities~\citep{saunshi2025reasoninglatentthoughtspower,yang2023looped}, its potential for high-fidelity visual generation remains largely untapped. From a practical standpoint, compared to the traditional models, Looped Transformers (a) are extremely parameter efficient and can perform significantly more compute (FLOPs) per parameter,  
(b) can have higher throughput by minimizing the ``memory wall'' bottleneck. By reusing a compact set of shared parameters across $L$ iterations, looped transformers achieve up to $L\times$ higher arithmetic intensity (FLOPs per byte transferred from memory) compared to standard transformers of equivalent depth. This avoids the cost of repeated transfers between different units of the accelerator (GPUs/TPUs) required in large transformers, and  (c) can exhibit robustness against overfitting in data-constrained settings.

\begin{figure}[t]
    \centering
    \resizebox{0.8\linewidth}{!}{
        \input{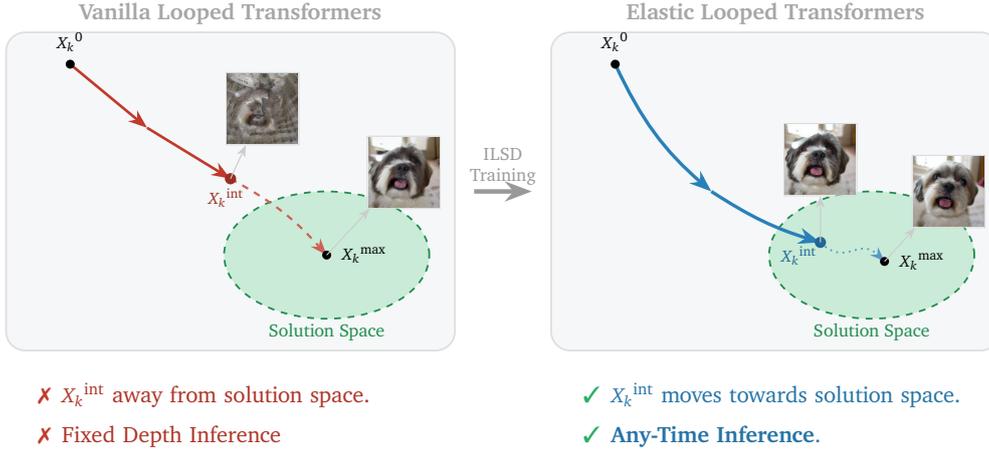}
    }
    \caption{\small \textbf{Latent Trajectories of Standard vs. Elastic Looped Transformers}.
    ${X_k}^{\text{int}}$ \& ${X_k}^{\text{max}}$ represent output of intermediate ($L_{\text{int}}$) \& final loops ($L_{\text{max}}$) respectively for $k^{th}$ generation sampling step. Unlike standard recurrent models (left) where only the final iteration ${X_k}^{\text{max}}$ reaches the solution space, our \distill\ training (right) guides intermediate states ${X_k}^{\text{int}}$ also toward the target space. This transformation shifts the model from a fixed-depth architecture to an Any-Time inference framework, supporting flexible computational budgets through early exits within a sampling step.}
    \label{fig:ilsd_trajectory}
\end{figure}

Despite the parameter efficiency of looped architectures, their training remains challenging because intermediate representations often remain uninterpretable until the final loop (see~\Cref{fig:ilsd_trajectory}). We address this by introducing \oursfull\ (\ours), a class of generative models designed for progressive refinement. Unlike traditional recurrent transformers, \ours\ is optimized to provide meaningful, high-quality synthesis even at intermediate repeats. This enables Any-Time (elastic) inference capability - where a single model can scale its compute based on available resources without sacrificing much on generation quality. We present a pictorial representation of our proposed method in~\Cref{fig:looping_main} and generation results for Any-Time inference capability in~\Cref{fig:gen_dit_ilsd}.

To achieve this functional flexibility across loops, we propose an \distillfull\ algorithm for training looped transformers. Rather than treating the loop as a fixed-depth process, our framework operates as a dual-path system: a teacher path executes the full loop count to provide a high-fidelity target, while a student path, defined strictly as a subset of the teacher’s trajectory, learns to produce comparable results with fewer iterations. Note that both the paths have the same parameter count. By framing the student process as a literal subset of the teacher’s forward propagation, we ensure there is no additional overhead during training. This approach forces the shared parameters to compress complex transformations into earlier loops. Consequently, the model does not just fix the output at the end. It learns an efficient, progressive refinement process that maintains the generation quality regardless of the exit point as motivated in~\Cref{fig:ilsd_trajectory}. Our contributions and findings are summarized as follows: 

{\flushleft \textbf{State-of-the-Art Parameter Efficiency}}: Through the reuse of block of transformer layers across loops, \ours\ achieves a competitive FID of \emph{2.0} on class-conditional ImageNet $256 \times 256$ and FVD of  \emph{72.8} on class-conditional UCF-101. This represents a \emph{4}$\times$ reduction in parameters compared to baselines MaskGIT~\citep{chang2022maskgit} (image generation) and MAGVIT~\citep{yu2023magvit} (video generation), while matching or improving their performance under iso-inference-compute settings.

{\flushleft \textbf{Elastic/Any-Time Inference}}: By treating the looped block as an iterative refiner, our models enable Any-Time inference~\citep{zilberstein1996using}, allowing to traverse the pareto frontier of quality versus compute at test-time without retraining. This allows to serve diverse deployment tiers: from latency-critical on-device generation (few loops) to high-fidelity cloud rendering (more loops).

{\flushleft \textbf{Scalability}}: While model size remains a primary driver of quality, recursive looping provides a unique test-time compute lever that scales predictably across both Masked Generative Transformers~\citep{chang2022maskgit, yu2023magvit, yu2023language} and Diffusion Transformers~\citep{peebles2023scalablediffusionmodelstransformers}.

\section{Related Work}
{\flushleft \textbf{Recursive Architectures:}} The principle of recursively applying a shared block of parameters to enhance model efficiency is a well-established concept. This approach, often referred to as \textit{looping} has been shaped by Universal Transformers~\citep{dehghani2018universal}, which introduced the idea of iterating over a single transformer layer. Notably, there has been a resurgence of interest in this area.~\citet{saunshi2025reasoninglatentthoughtspower} demonstrated the power of looping for sophisticated reasoning tasks.~\citet{gatmiry2024loopedtransformerslearnimplement} showed that Looped Transformers can learn to implement multi-step gradient descent for in-context learning, providing a deeper understanding of their capabilities.
Mixture-of-Recursions~\citep{bae2025mixtureofrecursionslearningdynamicrecursive} further explored input-dependent dynamic and adaptive depth in looped transformers. Mixture-of-Recursions-VIT~\citep{li2025morvitefficientvisiontransformer} extends it for image understanding.~\citet{fan2024looped} utilized looping for length generalization. Geiping et al.~\citet{geiping2025scalingtesttimecomputelatent} demonstrated that scaling test-time compute via recurrent depth allows language models to perform complex latent reasoning.

Deep Equilibrium Models (DEQs)~\citep{bai2019deepequilibriummodels, pokle2022deepequilibriumapproachesdiffusion, geng2023onestepdiffusiondistillationdeep, mccallum2025reversibledeepequilibriummodels, anil2022pathindependentequilibriummodels, gabor2024positiveconcavedeepequilibrium}, instead of unrolling a weight-tied layer for a fixed number of iterations, define the output as the fixed point of a non-linear transformation. Unlike DEQs that rely on black-box solvers for an analytical fixed point, our \ours\ framework explicitly optimizes unrolled intermediate states via \distillfull, retaining the flexibility of Any-Time inference without requiring the network to reach a strict analytical fixed point.

{\flushleft \textbf{Parameter-Efficient Visual Generation:}}
Standard efficiency techniques~\citep{menghani2023efficient} that work across deep learning models also help in speeding up visual generation models. MobileDiffusion~\citep{zhao2024mobilediffusioninstanttexttoimagegeneration} prunes redundant residual blocks and replaces standard layers with separable convolutions in UNet to get an optimized architecture ($\sim$400M) for on-device visual generation. EdgeFusion~\citep{castells2024edgefusionondevicetexttoimagegeneration} employs BK-SDM, a lightweight Stable Diffusion variant, and refines the step distillation process of Latent Consistency Model. MaGNeTS~\citep{goyal2025maskedgenerativenestedtransformers} trains a family of nested transformers~\citep{kudugunta2023matformer, kusupati2022matryoshka} with schedules of model sizes over the generation process, without increasing the parameter count.

{\flushleft \textbf{Elastic Visual Generation:}}
The paradigm of Any-Time or elastic generation focuses on decoupling model's parameter count from its computational depth. E-DiT~\citep{wang2026elasticdiffusiontransformer} introduces adaptive block skipping and MLP width reduction, allowing a single model to traverse varying computational budgets without retraining. In the context of visual reasoning, LoopViT ~\citep{shu2026loopvitscalingvisualarc} uses a weight-tied recursive architecture, employing a parameter-free dynamic exit mechanism to halt inference based on uncertainty of prediction. EvoSearch~\citep{he2025scalingimagevideogeneration} proposes a search-based strategy that optimizes sampling trajectories at inference time. Unlike these methods that rely on architectural skipping or external search, our \ours\ framework equipped with \distillfull\ algorithm directly regularizes the recursive process, ensuring stability across the entire loop spectrum. Methods like ALIT~\citep{duggal2024adaptivelengthimagetokenization}, FlexTok~\citep{bachmann2025flextokresamplingimages1d}, One-D-Piece~\citep{miwa2025onedpieceimagetokenizermeets}, ElasticTok~\citep{yan2025elastictokadaptivetokenizationimage}, CAT~\citep{shen2025catcontentadaptiveimagetokenization}, and DC-DiT~\citep{haridas2026dc} use tail dropping to allow elasticity in token sequence length.

{\flushleft \textbf{Relation to Few-Step and Consistency Models}:}
Recent approaches such as Consistency Models~\citep{song2023consistencymodels} and progressive distillation~\citep{salimans2022progressivedistillationfastsampling} address inference efficiency by reducing the number of sampling steps (inter-step acceleration). \ours\ is fundamentally orthogonal: it reduces compute \emph{within} each sampling step by varying the loop count $L$ (intra-step acceleration). These two axes are complementary, we can combine \ours\ with few-step methods to achieve further efficiency gains. Moreover, unlike consistency models which require specialized training objectives and architectural constraints to enable variable-step inference, \ours's elastic capability emerges naturally from \distill\ applied to standard training objectives. We note that \ours\ is particularly compelling for one-step generative paradigms, where the loop count $L$ becomes the \emph{sole} lever for controlling the compute-quality trade-off at inference time.

\section{Preliminaries}
\label{sec:pre}
{\flushleft \textbf{Masked Generative Models}:}
Masked Generative Image Transformer (MaskGIT)~\citep{chang2022maskgit} introduced a novel approach to image generation that significantly differs from traditional autoregressive models. In autoregressive decoding, images are generated sequentially, one pixel/token at a time, following a raster scan order~\citep{esser2021taming,kondratyuk2023videopoet, wang2024parallelizedautoregressivevisualgeneration, li2024autoregressiveimagegenerationvector}. This sequential approach is computationally inefficient, as each token is conditioned only on the previously generated tokens, leading to a bottleneck in processing time. MaskGIT generates all tokens of an image simultaneously, while iteratively refining them. This method enables significant acceleration in the decoding process. The tokens are discrete and obtained using Vector Quantized (VQ) autoencoders, learned with self-reconstruction and photo-realism losses~\citep{yu2023magvit}. The iterative parallel decoding process is represented as:
\begin{equation}
    \mathbf{X}_{k} \leftarrow \mathrm{Mask \circ Sample}({M}({\mathbf{X}}_{k-1}, c), k)
    \label{eqn:maskgit}
\end{equation}
where $\mathbf{X} \in \mathbb{Z}^N_{\geq 0}$, are the input tokens, $N$ is the number of tokens, $k \in [1, K]$ denote the iteration number, with $K$ being the total number of iterations, $\mathbf{X}_0$ is either completely masked for full generation, and partially masked for conditional generation tasks like frame prediction, $c$ is the category of image/video under generation. The $\mathrm{Sample}$ function utilizes logits predicted by the model ${M(.)}$, introduces certain randomness, and sorts them by confidence, unmasking only top-p tokens while masking the rest. 

{\flushleft \textbf{Diffusion Models}:} 
Diffusion models~\citep{ho2020denoising, song2020score} generate data by learning to reverse a process that gradually corrupts a signal $\mathbf{X}_0$ into Gaussian noise $\mathbf{X}_T$ through a predefined noise schedule.  At its core, diffusion denoising/sampling proceeds iteratively as follows:
\begin{align*}
    \mathbf{X}_{t-1} \leftarrow \alpha_1(t) \cdot \mathbf{X}_{t} + \alpha_2(t) \cdot M(\mathbf{X}_{t}, t, c) + \alpha_3(t) \cdot \mathbf{z}
\end{align*}
where $\mathbf{X}_{t}$ is the (partially) denoised vector at time $t$, $\alpha_1(t), \alpha_2(t), \alpha_3(t) $ are time-dependent scalars defined by the noise schedule and $\mathbf{z}$ is a sample drawn from standard Normal distribution. The model $M$ takes as input $\mathbf{X}_{t}$ along with time $t$ and class label $c$. From the perspective of the forward diffusion process, the generative task can be thought of as predicting the noise added to a latent representation $\mathbf{X}_t$ at timestep $t$, conditioned on a class label $c$. While traditionally architectures based on U-Nets~\citep{ronneberger2015unetconvolutionalnetworksbiomedical} have been used, 
the Diffusion Transformer (DiT)~\citep{peebles2023scalablediffusionmodelstransformers} architecture shifts away from this design by treating image latents as sequences of tokens, and using transformer blocks for processing these tokens. Similar to MaskGIT, we explore replacing the typical DiT transformers with \oursfull\ for denoising at time $t$.

In summary, both masked generative models and diffusion models involve recursive refinement over multiple sampling steps, sharing parameters across sampling steps. Unlike standard transformers, our \ours\ framework aligns the architecture of model $M$ with the progressive refinement process by implementing $M$ as a recurrent, weight-shared transformer blocks. This allows the model to perform recursive refinement within each sampling step, providing a test-time compute lever to trade-off inference speed and generation quality.

\section{Method}
\label{sec:method}

{\flushleft \textbf{Looping Mechanism}:} Let the number of transformer layers to be looped be $N$ and number of loops per sampling step be $L$. The total effective depth for a single sampling or denoising step of the generation process is then $N \times L$.
Let $f_{\theta_i}(\mathbf{x})$ denote a single transformer layer with parameters $\theta_i$. We define a composite block $g_{\Theta}(x)$ consisting of $N$ unique transformer layers with parameters $\Theta = \{\theta_1, \theta_2, \dots, \theta_N\}$ as follows:
\begin{align*}
    g_{\Theta}(\mathbf{x}) = f_{\theta_N}(f_{\theta_{N-1}}(\cdots f_{\theta_1}(\mathbf{x})))
\end{align*}

\begin{figure}[t]
    \centering
    \resizebox{0.9\linewidth}{!}{
        \input{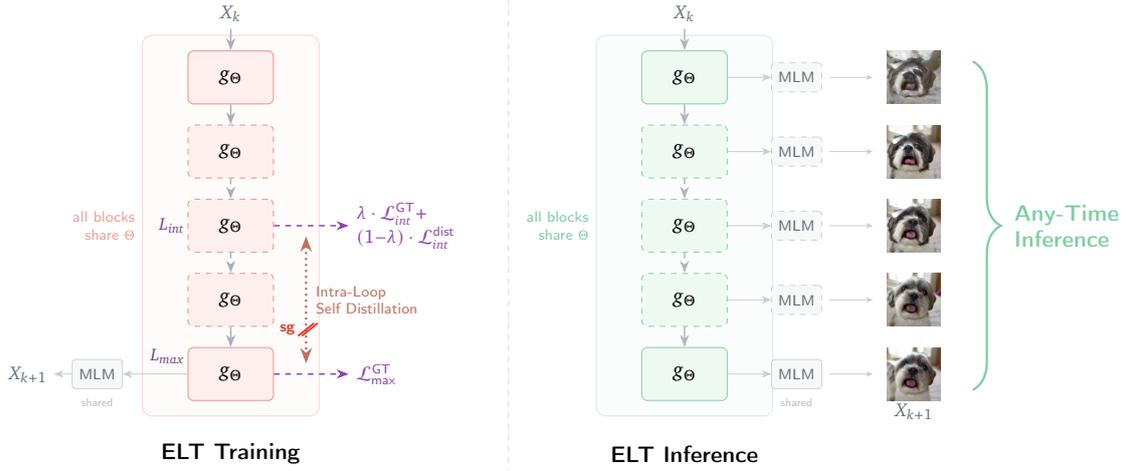}
    }
    \caption{\textbf{Overview of \ours\ framework}. Training (left) utilizes shared parameters $\Theta$ with recurrent loops and \distillfull\ to improve intermediate representations. This enables Any-Time inference (right), allowing the model to exit early and predict $X_{k+1}$ from any intermediate block via a shared MLM head. Note that $X_{k}$ represents the input for $k^{th}$ sampling iteration and output images in \ours\ Any-Time Inference (right) assume  $k = K$ (last sampling step). Refer~\Cref{sec:pre} for details of annotations.}
    \label{fig:looping_main}
    \vspace{-0.5cm}
\end{figure}

In a standard transformer model with total depth $\mathcal{D} = N \times L$, the effective transformation $F_{\mathcal{D} }(x)$ would require $\mathcal{D} $ sets of unique parameters. In \textit{looping}, we reuse the same block of parameters $\Theta$ for $L$ successive applications, resulting in only $N$ unique layers of parameters. The effective transformation for a $N \times L$ configuration is given by:
\begin{align*}
    F_{(N, L)} (\mathbf{x}) =   \underbrace{g_{\Theta}\left(g_{\Theta} (\cdots g_{\Theta}(\mathbf{x}))\right)}_{L \text{ loops}} \equiv g_{\Theta}^L(\mathbf{x})
\end{align*}

This looping architecture decouples physical model size from computational depth (see~\Cref{fig:looping_main} for a visual overview). The parameter count ($\Theta$) is constrained by the number of unique blocks ($N$), while representational capacity and depth ($\mathcal{D}$) scale with loop count ($L$). This setup provides two primary advantages: extreme parameter efficiency and high throughput. See~\Cref{sec:image_gen} for details.

{\flushleft \textbf{\distillfull}:}
In a standard weight-tied (looped) transformer, the model is typically optimized only for its final output after fixed $L_{\text{max}}$ iterations i.e. the loop count for which it is trained. However, this creates a ``black box'' internal trajectory where intermediate loops ($L < L_{\text{max}}$) may produce suboptimal representations until the final projection layer. By treating the looped block $g_{\Theta}$ as an iterative refiner, we can motivate a training objective that ensures the model remains useful at multiple depths. This not only encourages the model to learn a more robust, incremental transformation but also allows for elastic inference, where the model can be exited early with minimal performance drop. Towards this end, we propose \distillfull.

From a distillation perspective, \distill\ leverages the fact that a model with more loops ($L_{\text{max}}$) naturally possesses a more mature and refined representational space than its shallower version ($L_{\text{int}}$), even though they have the same unique parameters. By treating the full-depth model as an internal teacher, we provide a high-fidelity, low-variance signal for the shallower student to follow (see~\Cref{fig:looping_main}). This forces the shared parameters $\Theta$ to compress complex transformations into fewer steps, effectively distilling the knowledge of the deep model into the early stages of the computation.

{\flushleft \textbf{Stochastic Student Sampling ($S^3$)}:}
We train with a fixed number of loops, $L_{\text{max}}$, for a block of $N$ layers with unique parameters. During each training step, we treat the model as a dual-path system sharing a single set of parameters $\Theta$. We define a teacher path that executes the full $L_{\text{max}}$ loops and a stochastic student path that exits early at an intermediate loop $L_{\text{int}}$. The intermediate path receives supervision from the ground truth labels as well as the online teacher with maximum training loop count $L_{\text{max}}$.

We denote the training loss for the output of a configuration $N \times L$ as $\mathcal{L}_{\Theta}(F_{(N, L)}(\mathbf{x}), \mathbf{y})$ where $\mathbf{y}$ denote the ground-truth for a certain input masked / denoise level. At every training iteration, we randomly sample an intermediate loop count $L_{\text{int}}$ from uniform distribution such that $L_{\text{min}} \leq L_{\text{int}} < L_{\text{max}}$. Note that $L_{\text{min}}$ is just used for constraining the student sampling distribution.
The effective joint loss, $\mathcal{L}^{\text{\distill}}_{\Theta}$, is computed as:

\begin{align*}
\mathcal{L}^{\text{\distill}}_{\Theta} = & \mathcal{L}^\text{GT}(F_{(N, L_{\text{max}})}(\mathbf{x}), \mathbf{y}) && \text{\small (1) Ground-truth for teacher} \\
+ &\lambda \mathcal{L}^\text{GT}(F_{(N, L_{\text{int}})}(\mathbf{x}),\mathbf{y}) && \text{\small (2) Ground-truth for student} \\
+ & (1 - \lambda) \mathcal{L}^{\text{dist}}(F_{(N, L_{\text{int}})}(\mathbf{x}), \text{sg}(F_{(N, L_{\text{max}})}(\mathbf{x}))) && \text{\small (3) Intra-Loop Self Distillation} \\
\text{with} \quad &L_\text{int} \sim \mathcal{U}(L_{\text{min}}, L_{\text{max}})
&& \text{\small  Stochastic Student Sampling}
\end{align*}
\noindent
where sg is stop-grad for teacher ($L_{\text{max}}$) in \distill, $\lambda$ is hyperparameter controlling the weight between the ground truth and distillation losses. We introduce a curriculum for $\lambda$ and linearly decay it from $1$ to $0$ as training progresses. This initially anchors the student to reliable ground-truth labels while the teacher is still untrained and gradually shift to mimicking the teacher's predictions once they have matured. We found this linear schedule to be effective across all our experiments and did not observe sensitivity to the decay rate, as long as the transition is gradual.

{\flushleft \textbf{Loss Formulation}:}
The exact forms of $\mathcal{L}^\text{GT}$ and $\mathcal{L}^\text{dist}$ depend on the algorithm used for training. For masked generative models with discrete tokens, we use the cross-entropy loss for both ground-truth and distillation:
\begin{align*}
\mathcal{L}^\text{GT} &= -\sum_{i \in Mask} \log P_{(N, L_{\text{int}})}(y_i \mid \mathbf{x}_{mask}) \\
\mathcal{L}^\text{dist} &= -\sum_{i \in Mask} \sum_{v \in \mathcal{V}} P_{(N, L_{\text{max}})}(v \mid \mathbf{x}_{mask}) \log P_{(N, L_{\text{int}})}(v \mid \mathbf{x}_{mask})
\end{align*}
where $\mathbf{x}_{mask}$ is the masked input, $y = \{y_i\}_{i \in \text{Mask}}$ represents the ground-truth tokens for the masked positions, and $\mathcal{V}$ is the full vocabulary of the tokenizer. For diffusion, we use the sigmoid-weighted Mean Squared Error (MSE) for both ground-truth and distillation losses. Let $\mathbf{x}_t$ be the noised version of ground truth latent $\mathbf{x}_0$, the ground-truth loss is:
\begin{align*}
    \mathcal{L}^{\text{GT}}= w(t) \norm{F_{(N, L)}(\mathbf{x}_t) - \mathbf{x}_0}_2^2
\end{align*}
where $L$ is $L_{\text{int}}$ for student and $L_{\text{max}}$ for teacher. $w(t)$ is a time-dependent sigmoid weighting term~\citep{hoogeboom2025simpler}. The distillation loss is:
\begin{align*}
    \mathcal{L}^{\text{dist}}= w(t) \norm{F_{(N, L_{\text{max}})}(\mathbf{x}_t) - F_{(N, L_{\text{int}})}(\mathbf{x}_t)}_2^2
\end{align*}

Note that gradients from both computational paths, corresponding to $L_{\text{int}}$ and $L_{\text{max}}$, update the single, shared set of block parameters $\Theta$. This joint optimization provides a significantly richer training signal; the shared block $g_{\Theta}$ is forced to learn a transformation that is not only effective at $L_{\text{int}}$ loops but remains incrementally useful for the subsequent iterations up to $L_{\text{max}}$ loops. This constraint prevents the model from learning a \textit{shortcut} that might minimize loss at a specific depth but fail when composed further. Consequently, the shared block generalizes better to lower depths, leading to higher performance even with fewer loops. It is interesting to note that unlike traditional distillation, where we have to forward propagate through both the student and teacher models separately, in the proposed way of \distillfull, the training overhead is minimal, as the computation of $F_{(N, L_{\text{int}})}(\mathbf{x})$ is a strictly required intermediate step for computing $F_{(N, L_{\text{max}})}(\mathbf{x})$ i.e. the student trajectory ($L_{\text{int}}$) is a strict prefix of the teacher trajectory ($L_{\text{max}}$). Refer \Cref{alg:elt_train} and \Cref{alg:elt_infer} for details of our training and inference algorithms respectively.

\section{Experiments and Results}

We conduct extensive experiments using masked generative transformers and diffusion transformers to demonstrate the generality and efficacy of our approach on class-conditional image generation. We also experiment with class-conditional video generation using masked generative transformers. We first detail our experimental setup and then present the results.

\begin{table}[hbt!]
\centering
\caption{\small \textbf{ Class-conditional Image Generation} on ImageNet $256\times256$. “\# steps” refers to the number of neural network runs. $^{\Delta}$ denotes values taken from prior publications. $^{*}$ indicates usage of extra training data. $^g$ denotes use of classifier-free guidance \citep{ho2022classifierfreediffusionguidance}. Note that $L$ in ($N \times L$) notation is inference loop count per sampling step. }
\vspace{-2mm}
\resizebox{0.8\columnwidth}{!}{%
\begin{tabular}{lccccccc}
    \toprule
    Model & AR & FID $\downarrow$ & IS $\uparrow$ & \# params & \# steps & \# Gflops\\
    \midrule
    BigGAN-deep$^{\Delta}$ \citep{brock2018large}&  & 7.0 & 171.4 & 160M & 1 & - \\
    StyleGAN-XL$^{\Delta g}$ \citep{sauer2022styleganxlscalingstyleganlarge}&  & 2.3 & 265.1 & 166M & 1 & - \\
    \midrule    
    Improved DDPM$^{\Delta}$ \citep{nichol2021improved}&  & 12.3 & - & 280M & 250 & $>$150k\\
    ADM + Upsample$^{\Delta g}$ \citep{dhariwal2021diffusion}&  & 3.9 & 215.8 & 554M & 250 &  371k\\
    LDM-4$^{\Delta g*}$  \citep{ldm}& & 3.6 & 247.7 & 400M & 250 &  51.5k\\
    DiT-XL/2$^{\Delta g*}$ \citep{peebles2023scalablediffusionmodelstransformers}& & 2.3 & 278.2 & 675M & 250 & 59.5k \\ 
    MDT$^{\Delta g* }$ \citep{gao2023masked}&  & 1.8 & 283.0 & 676M & 250 & $>$59k \\ 
    MaskDiT$^{\Delta g* }$ \citep{zheng2023fast}&  & 2.3 & 276.6 & 736M & 250 & $>$28k   \\
    CDM$^{\Delta}$ \citep{ho2022cascaded}&  & 4.9 & 158.7 & - & 8100 & -  \\
    RIN$^{\Delta}$ \citep{jabri2022scalable}&  & 3.4 & 182.0 & 410M & 1000 & 334k  \\
    Simple Diffusion$^{\Delta g}$ \citep{hoogeboom2023simple}&  & 2.4 & 256.3 & 2B & 512 & - \\
    VDM++$^{\Delta g}$  \citep{kingma2023understanding}& & 2.1 & 267.7 & 2B & 512 & - \\
    EDiff$^{\Delta g}$  \citep{hang2024efficientdiffusiontrainingminsnr}&  & 2.1 & - & 450M & 50 & 119k \\
    LPDM-ADM$^{\Delta g}$  \citep{wang2023patchdiffusionfasterdataefficient}&  & 2.7 & - & - & 50 & 7.8k \\
    MAR$^{\Delta g}$  \citep{li2024autoregressiveimagegenerationvector}& $\checkmark$ & 1.8 & 296.0 & 479M & 128 & - \\
    \midrule
    VQVAE-2$^{\Delta}$ \citep{razavi2019generating}& $\checkmark$ & 31.1 & $\sim$45 & 13.5B & 5120 & -\\
    VQGAN$^{\Delta}$ \citep{esser2021taming}& $\checkmark$ & 15.8 & 78.3 & 1.4B & 256 & -\\
    MaskGIT$^{\Delta}$\citep{chang2022maskgit}& & 6.2 & 182.1 & 227M & 8 & 647 \\
    Mo-VQGAN$^{\Delta}$ \citep{zheng2022movqmodulatingquantizedvectors}& & 7.2 & 130.1 & 389M & 12 & $\sim$1k\\
    MaskBit$^{\Delta g}$ \cite{weber2024maskbitembeddingfreeimagegeneration}&  & 1.7 & 341.8 & 305M & 64 & 10.3k\\
    PAR-$4\times^{\Delta}$ \cite{wang2024parallelizedautoregressivevisualgeneration}&$\checkmark$ & 3.8 & 218.9 & 343M & 147 & -\\
    PAR-$16\times^{\Delta}$ \cite{wang2024parallelizedautoregressivevisualgeneration}&$\checkmark$ & 2.9 & 262.5 & 3.1B & 51 & -\\
    \midrule
    MaskGIT-L$^g$ & & 2.1 & 270.1 & 303M & 24 & 3.7k \\
    MaskGIT-XL$^g$ & & 2.0 & 294.8 & 446M & 24 & 3.9k \\
    \midrule
    \textbf{\ours-L} ($\bm{8N \times 3L}$)& & 2.2 & 254.3 & \textbf{101M} & 24 & 3.7k \\
    \textbf{\ours-L} ($\bm{12N \times 2L}$)& & 2.1 & 281.8 & \textbf{152M} & 24 & 3.7k \\
    \textbf{\ours-XL} ($\bm{7N \times 4L}$)& & 2.0 & 266.1 & \textbf{111M} & 24 & 3.9k \\
    \bottomrule
\end{tabular}
}

\label{tab:img_results}
\end{table}

\subsection{Experimental Setup}

\label{sec:exp_setup}

{\flushleft \textbf{Datasets}:}
We experiment on ImageNet $256\times256$~\citep{deng2009imagenet} for image generation and UCF-101~\citep{soomro2012ucf101dataset101human} for class-conditional video generation.

{\flushleft \textbf{Implementation Details}:} \textbf{(i) Masked Generative Transformers}: We use pretrained tokenizers from MaskGIT~\citep{chang2022maskgit} (images) and MAGVIT~\citep{yu2023magvit} (videos) with a codebook size of 1024 tokens. Image models are trained at $256\times256$ resolution, compressed to $16\times16$ discrete tokens with an embedding dimension of 1024. Video models are trained for $16 \times 128 \times 128$ sequences, compressed to $4 \times 16 \times 16$ tokens. Following MaskGIT, we adopt the BERT~\citep{devlin2019bertpretrainingdeepbidirectional} architecture as the transformer backbone and
perform experiments at several model scales to understand the scaling behaviors of ELT.  We train all models for 270 epochs unless otherwise specified. We employ a cosine schedule for unmasking tokens during inference. For image generation, we use classifier-free guidance by dropping class condition labels for $10\%$ of the training batches. \textbf{(ii) Diffusion Transformers:}  We use a pretrained Stable Diffusion v1.4 VAE~\citep{ldm} model to map $256 \times 256$ images into a continuous $32 \times 32 \times 4$ latent space (8× spatial downsampling). We train a DDPM-style diffusion model which operates on these latents using a DiT architecture. We employ a shifted cosine noise schedule and sigmoid-weighted MSE loss for training~\citep{hoogeboom2025simpler}. Models are trained for 500K steps with a batch size of $512$ using Adam. Unless mentioned otherwise, sampling uses 512-step DDPM with classifier-free guidance scale $3.0$. See Appendix for more details.

{\flushleft \textbf{Evaluation Metrics and Efficiency}}:
To evaluate the quality of synthesized content, we employ Fréchet Inception Distance (FID)~\citep{heusel2017gans, dhariwal2021diffusion} and Inception Score (IS)~\citep{salimans2016improvedtechniquestraininggans} for image generation tasks, and Fréchet Video Distance (FVD)~\citep{unterthiner2019fvd} for video generation. Beyond generative quality, we evaluate model efficiency using inference-time GFLOPs and throughput (samples generated per second). In the proposed $N \times L$ design space, for a fixed model scale and block size $N$, both computational complexity and latency scale linearly with the number of loops $L$. This relationship allows us to precisely navigate the trade-off between generation quality and inference speed by modulating the loop count.

\begin{figure}[t!]
    \centering
    \includegraphics[width=0.9\textwidth]{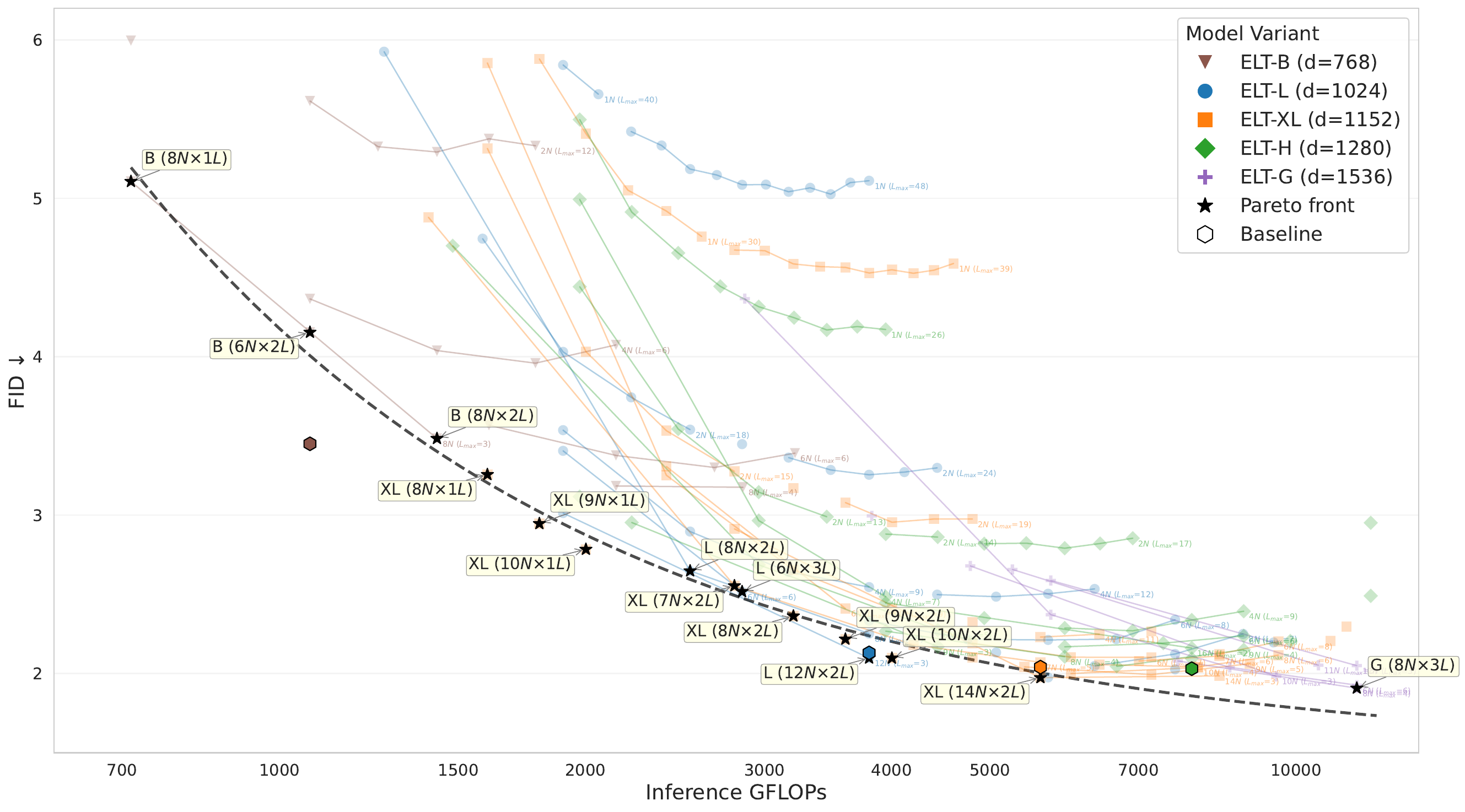}
    \vspace{-1mm}
    \caption{\small \textbf{Pareto front of FID vs. Inference GFLOPs}. The black curve denotes the fit (FID = $1922.5 \cdot G^{-0.95} + 1.48$) over pareto-optimal configurations, representing the best achievable FID for a given computational budget. Points are labeled as $N \text{ layers} \times L \text{ loops}$. Results demonstrate that \ours\ scales as effectively as the baseline while remaining significantly more parameter-efficient. Scaling both the model dimension ($d$) and the number of loops ($L$) follows a predictable efficiency frontier, where larger models with fewer loops often compete with smaller models with more loops at specific target GFLOPs. Trends across faded points show scaling along number of loops $L$ in inference, from a single run trained with a certain $L_{max}$ loops. Refer~\Cref{tab:model_scaling} for the exact baseline configuration, including the specific number of layers for each comparison point.
    } 
    \label{fig:fid_vs_gflops_imagenet}
\end{figure}

\begin{figure}[ht!]
    \centering
    \vspace{-1mm}
    \includegraphics[width=\textwidth]{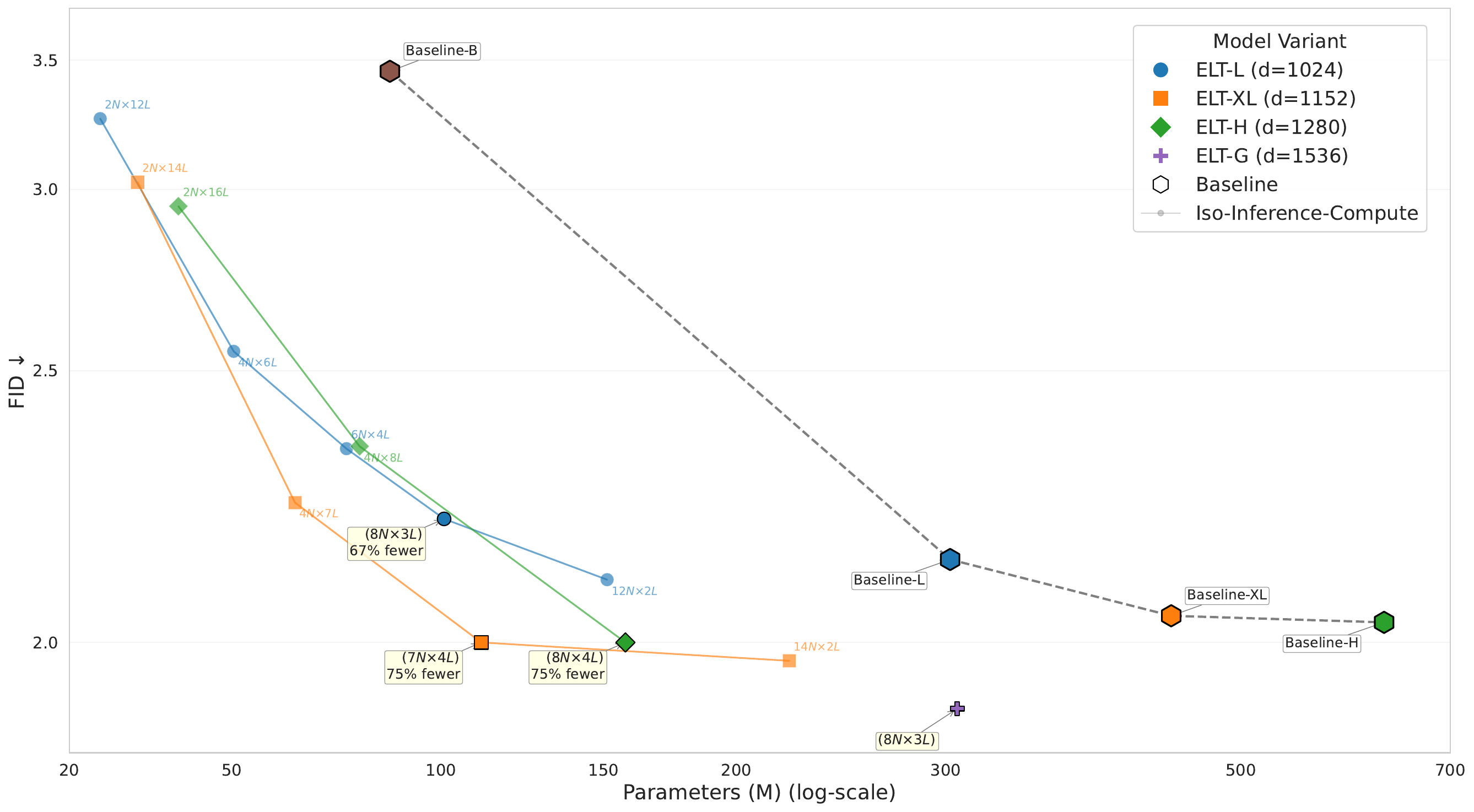}
    \caption{\small \textbf{Scaling with Parameters}. We show the best-achievable FID as a function of model parameters (log scale). Each series corresponds to iso-inference-compute configuration for a specific model width ($d$). Points are labeled as $N \ \text{layers} \times L \ \text{loops}$. Results show that increasing parameter count via model width yields superior FID. Proposed ELT mechanism shows that even parameter-constrained models can achieve similar performance through recursive depth. Each \ours\ point in the figure is best inference configuration ($N \times L$) chosen from its corresponding training run ($N \times L_{\text{max}}$). Refer~\Cref{tab:model_scaling} for the exact baseline configuration, including the specific number of layers for each comparison point.
    }
    \label{fig:fid_vs_params_imagenet}
\end{figure}

\subsection{Image Generation}
\label{sec:image_gen}

{\flushleft \textbf{Comparison with Baselines}}:  We present the results for $256 \times 256$ image generation on ImageNet-1k in~\Cref{tab:img_results}. Despite using \textbf{4}$\times$ less parameters, ELT-XL achieves same FID of 2.0 as MaskGIT-XL, which is the base setup for ELT. As shown in recent literature, using a superior tokenizer~\citep{zhao2024imagevideotokenizationbinary, yu2023language, weber2024maskbitembeddingfreeimagegeneration} or optimized training \& inference configurations~\citep{ni2024revisitingnonautoregressivetransformersefficient, ni2024enatrethinkingspatialtemporalinteractions} can further boost \ours's performance. 

\begin{figure}[ht]
    \centering
    
    \begin{subfigure}[t]{0.49\textwidth}
    \captionsetup{justification=centering}
        \centering
        \includegraphics[height=45mm]{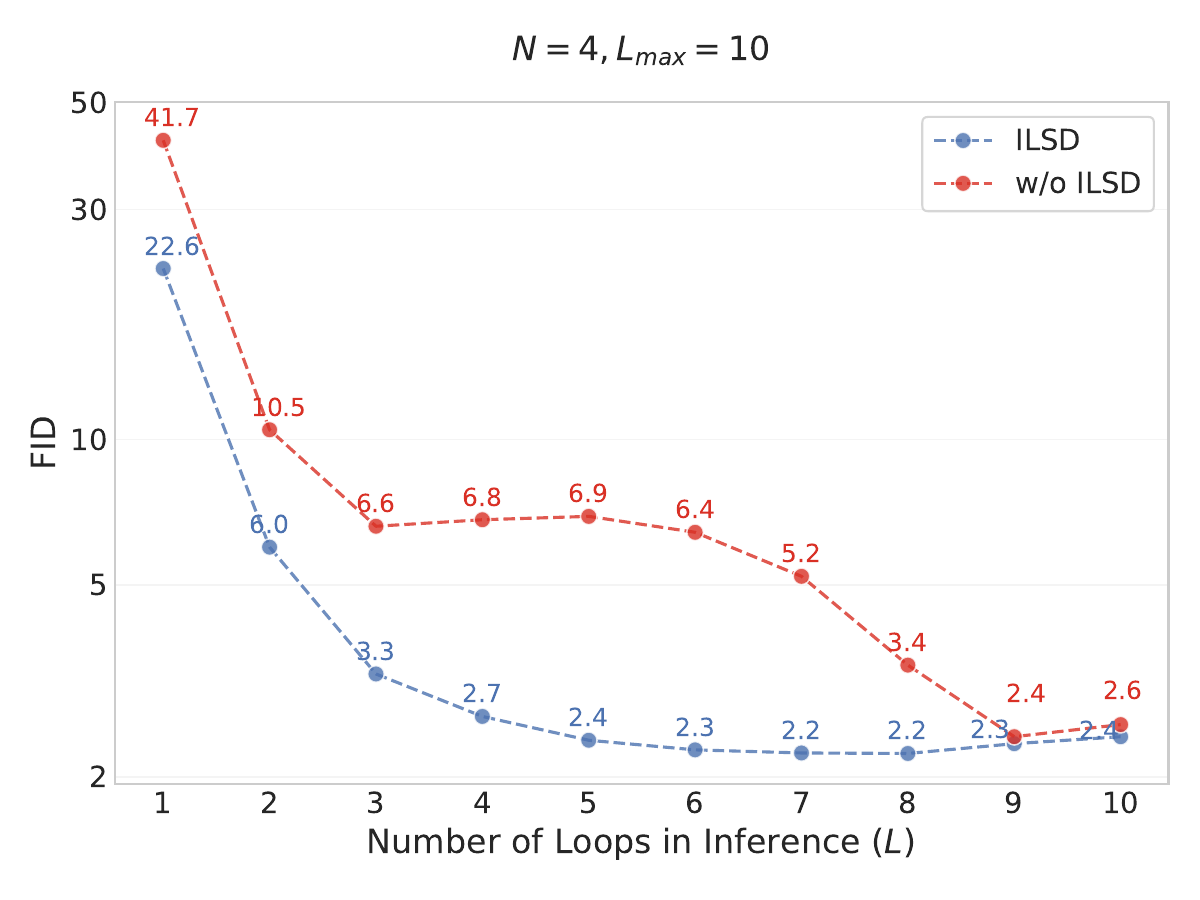}
        \caption{Masked Generative Models}
        \label{fig:ilsd_effect_maskgit}
    \end{subfigure}
    \hfill 
    \begin{subfigure}[t]{0.5\textwidth}
    \captionsetup{justification=centering}
        \centering
        \includegraphics[height=45mm]{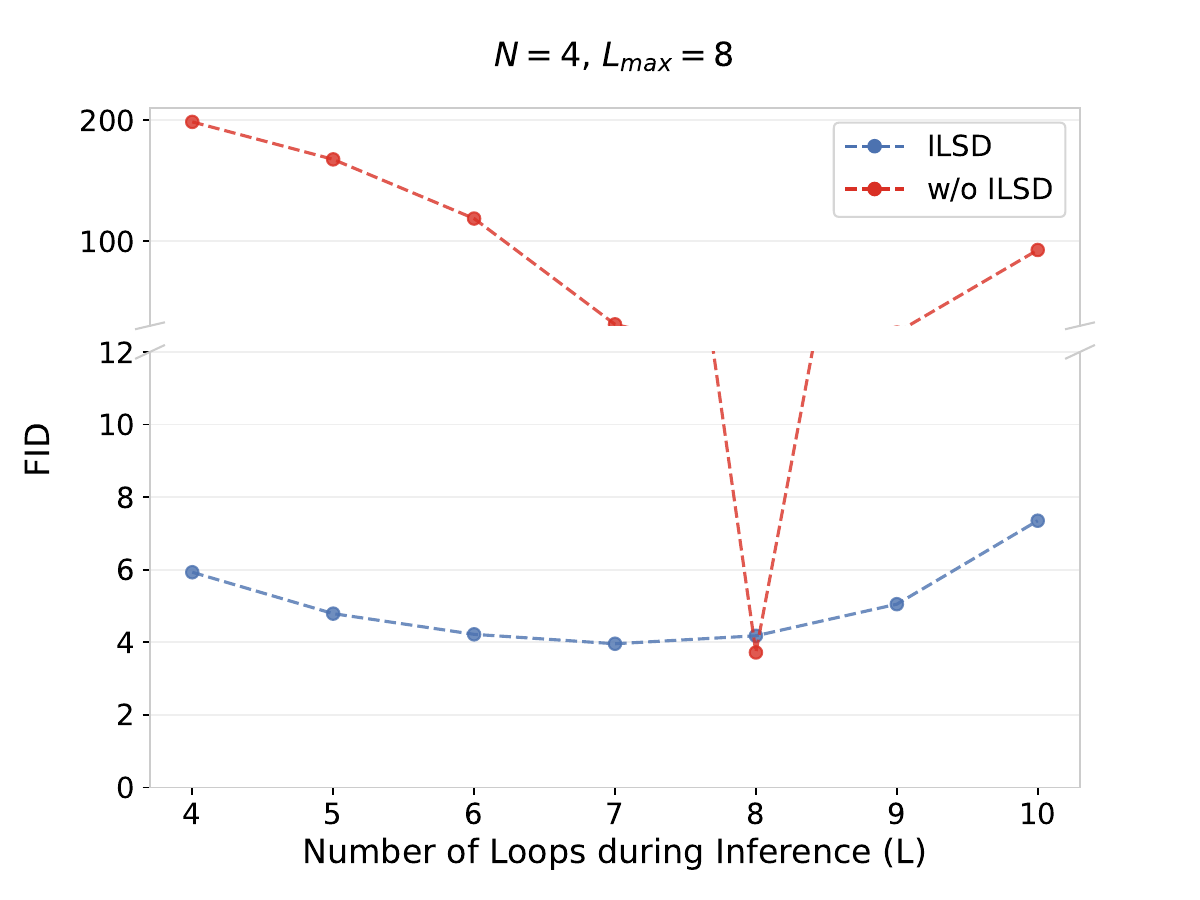}
        \caption{Diffusion Transformers}
        \label{fig:ilsd_dit}
    \end{subfigure}
    \caption{\small \textbf{Impact of \distill\ across generation processes.} \distill\ improves performance for all $L$ in inference especially when $L \neq L_{max}$. For w/o ILSD, $L_{max}$ here refers to the number of loops the model is trained with. Results are on ImageNet $256 \times 256$.}
    \label{fig:distill_impact_combined}
\end{figure}

We also present comparisons of \ours\ using diffusion models in~\Cref{tab:dit}. Notably, \ours\ outperforms the baseline model with 32 layers (FID 3.43) using iso-inference-compute $8N \times 4L$ (FID 3.16) and $16N \times 2L$ (FID 2.83) settings, achieving $\textbf{4} \times$ and $\textbf{2} \times$ reduction in parameter count respectively. 
The $1N \times 32L$ configuration (FID 10.30) reveals that a single unique transformer layer, despite 32 effective passes, lacks the representational capacity for high-fidelity generation, highlighting that a minimum block size $N$ is necessary to provide sufficient architectural expressiveness within each iteration. While vanilla looping (without \distill) gives competitive performance when running inference with same loops as training ($L=L_{max}$), performance degrades drastically for lower number of loops which is mitigated by \distill\ as show in~\Cref{fig:ilsd_dit}.

\begin{figure}[h!]
    \centering
    \includegraphics[width=\textwidth]{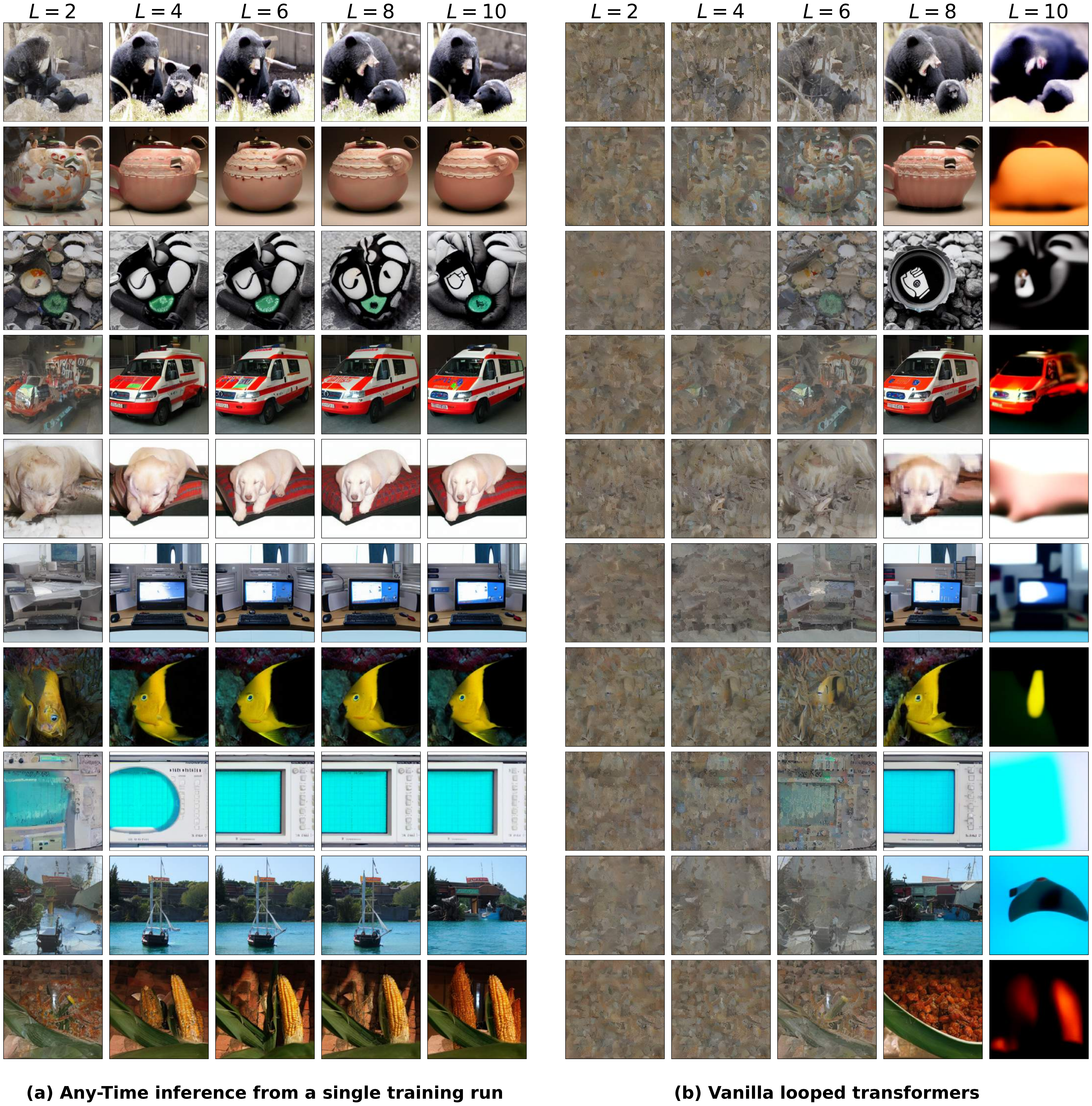}
      \caption{\small \textbf{Class-conditional Image Generation on ImageNet $\bm{256 \times 256}$}. Comparing our proposed \oursfull\ (left) against vanilla looped transformers (right), which simply repeat transformer layers during a forward pass of diffusion. As shown, standard looping only produces a coherent image when the inference loop count exactly matches its training setup ($L=8$), with quality severely degrading at all other values. In contrast, our proposed \ours\ enhances vanilla looping with Self Distillation, maintaining a high-fidelity generation across several evaluated compute budgets.}
    \label{fig:dit_qual_supp1}
\end{figure}

\begin{figure}[h!]
    \centering
    \includegraphics[width=\textwidth]{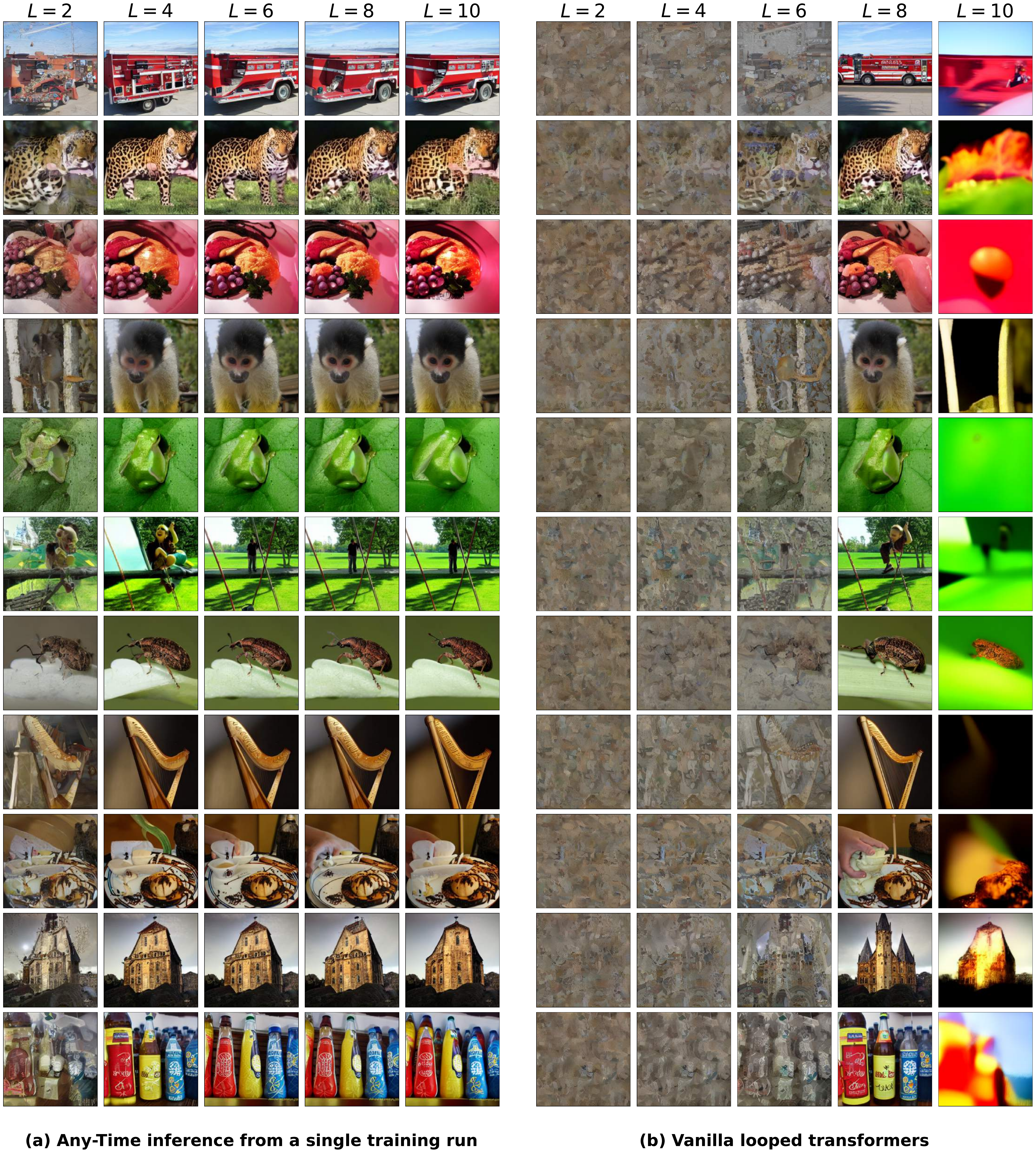}
      \caption{\small \textbf{Class-conditional Image Generation on ImageNet $\bm{256 \times 256}$}. Comparing our proposed \oursfull\ (left) against vanilla looped transformers (right), which simply repeat transformer layers during a forward pass of diffusion. As shown, standard looping only produces a coherent image when the inference loop count exactly matches its training setup ($L=8$), with quality severely degrading at all other values. In contrast, our proposed \ours\ enhances vanilla looping with Self Distillation, maintaining a high-fidelity generation across several evaluated compute budgets.}
    \label{fig:dit_qual_supp2}
\end{figure}

\begin{figure}[ht!]
    \centering
    \includegraphics[width=0.9\textwidth]{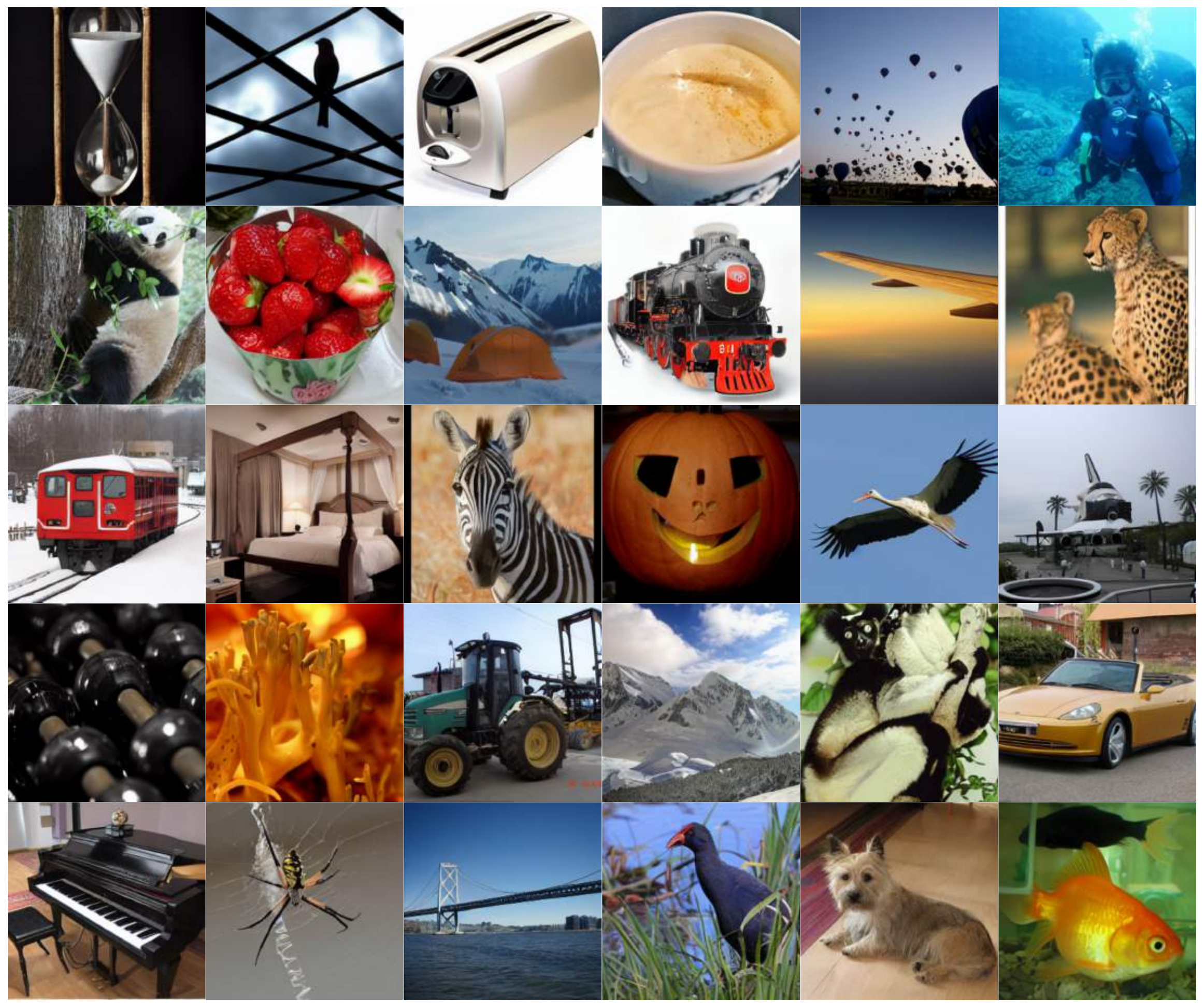}
    \caption{\small \textbf{Class-conditional Image Generation on ImageNet $\bm{256 \times 256}$}. Qualitative results for \ours\ in diffusion framework, $16N \times 2L$ inference configuration \& trained with $L_{\text{max}}=2$.
    }
    \label{fig:dit_final_gen_qual_supp}
\end{figure}

\begin{figure}[h!]
    \centering
    \includegraphics[width=0.9\textwidth]{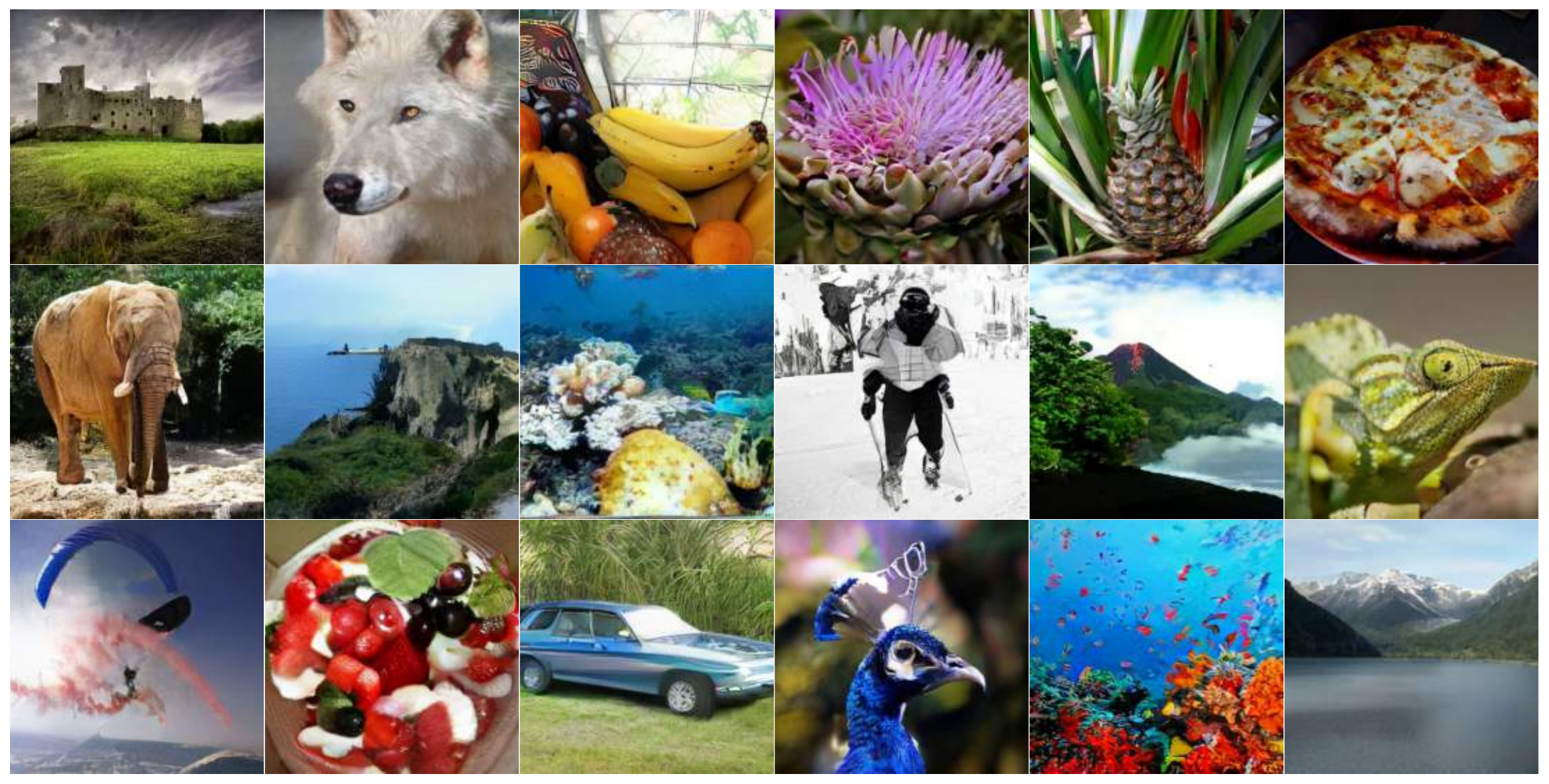}
    \caption{\small \textbf{Class-conditional Image Generation on ImageNet $\bm{256 \times 256}$}. Qualitative results for \ours\ in masked generative framework. Model scale: \ours-G, $8N \times 3L$ inference config \& trained with $L_{\text{max}}=4$ (FID=1.9).
    }
    \label{fig:qual_elt_maskgit}
\end{figure}

{\flushleft \textbf{Qualitative Results}}: \Cref{fig:gen_dit_ilsd}, \Cref{fig:dit_qual_supp1}, and \Cref{fig:dit_qual_supp2} compare \ours\ against vanilla looped transformers within a diffusion framework. It is clear that \ours\ unlocks Any-Time inference capability providing dynamic trade-off between generation quality and inference speed. \Cref{fig:dit_final_gen_qual_supp} and \Cref{fig:qual_elt_maskgit}  visualize the generation results of \ours\ in diffusion and masked generative framework respectively. 

{\flushleft \textbf{Scaling Inference GFLOPs and Pareto Front}:}
~\Cref{fig:fid_vs_gflops_imagenet} illustrates the trade-off between generation quality (FID) and inference compute (GFLOPs). The pareto front (black curve) reveals that while increasing the loop count ($L$) consistently improves FID (faded points) for a fixed unique layers ($N$), the gains eventually diminish, where  transitioning to the next architecture scale becomes more performant than over-looping smaller models. Crucially, ELT allows for Any-Time inference, we can traverse the pareto curve at test-time by simply adjusting $L$ to meet specific hardware constraints without retraining.

\begin{figure}[t!]
    \centering
    
    \begin{minipage}[c]{0.48\textwidth}
        \centering
        \captionof{table}{\small \textbf{Class-conditional Image Generation} on ImageNet $256\times256$ using DiT. We report $ N \times L$ used for inference in parenthesis in Model column.}
        \resizebox{0.8\columnwidth}{!}{
    \begin{tabular}{lccc}
        \toprule
        {Model} & {FID $\downarrow$} & {\# params} & $\mathcal{D}$\\
        \midrule
        DiT - 16 layers & $3.87$  & 1.1B & 16 \\
        DiT - 32 layers &  $3.43$  & 2.1B & 32 \\
        \midrule   
        \ours\ ($ 1N \times 32L$)& $10.30$ &69M & 32 \\
        \ours\ ($4N \times 8L$)&  $3.96$  & 271M & 32\\
        \textbf{\ours}\ ($\bm{8N \times 4L}$)&  $\textbf{3.16}$ & \textbf{539M}& 32\\
        \textbf{\ours}\ ($\bm{16N \times 2L}$)&  $\textbf{2.83}$ &\textbf{1.1B}& 32\\
        \bottomrule
    \end{tabular}
}
        \vspace{2mm}
        \label{tab:dit}
    \end{minipage}
    \hfill
    \begin{minipage}[c]{0.48\textwidth}
        \centering
    \captionof{table}{\small \textbf{Throughput gains for \ours}. We report throughput (images/sec) ratio (\ours\ /\ Baseline) on a Google Cloud TPU v6e~\citep{googleclouduv6e} with $1\times1$ topology and inference batch size of 8. The speedup arises from \ours's compact shared parameters fitting on-chip, reducing repeated HBM-to-SRAM transfers. \ours\ achieves a peak speedup of $\emph{3.5}\times$ for model scale H.}

    \resizebox{0.8\columnwidth}{!}{
    \begin{tabular}{lcc}
    \toprule
    \ours\ & $d_{model}$ & Throughput Ratio \\ 
    \midrule
    $6N \times 2L$ (B)  & 768  & 1.0 \\
    $8N \times 3L$ (L)  & 1024 & 2.9 \\
    $7N \times 4L$ (XL) & 1152 & 3.3 \\
    $8N \times 4L$ (H)  & 1280 & \textbf{3.5} \\
    \bottomrule
    \end{tabular}
    }
    
    \label{tab:thru}
    \end{minipage}

\end{figure}

{\flushleft \textbf{Scaling Parameters}:} We further investigate the relationship between model capacity and performance in~\Cref{fig:fid_vs_params_imagenet}. By plotting the best FID achieved at each parameter budget, we observe a clear power-law scaling trend across all model widths. While increasing the number of unique layers ($N$) reduces FID, the gains are most pronounced when accompanied by an increase in model width ($d$). Specifically, the $d=1536$ configuration ($G$) achieves the lowest overall FID with only $8$ unique layers, while the full $G$ model has $48$ unique layers. However, for on-device budgets, smaller looped variants (L and XL) remain highly efficient, providing a flexible scale-to-performance ratio that is critical for resource-constrained visual generation.

\begin{figure}[htp]
    \centering
    
    \begin{subfigure}[t]{0.48\textwidth}
        \centering
        \includegraphics[height=45mm]{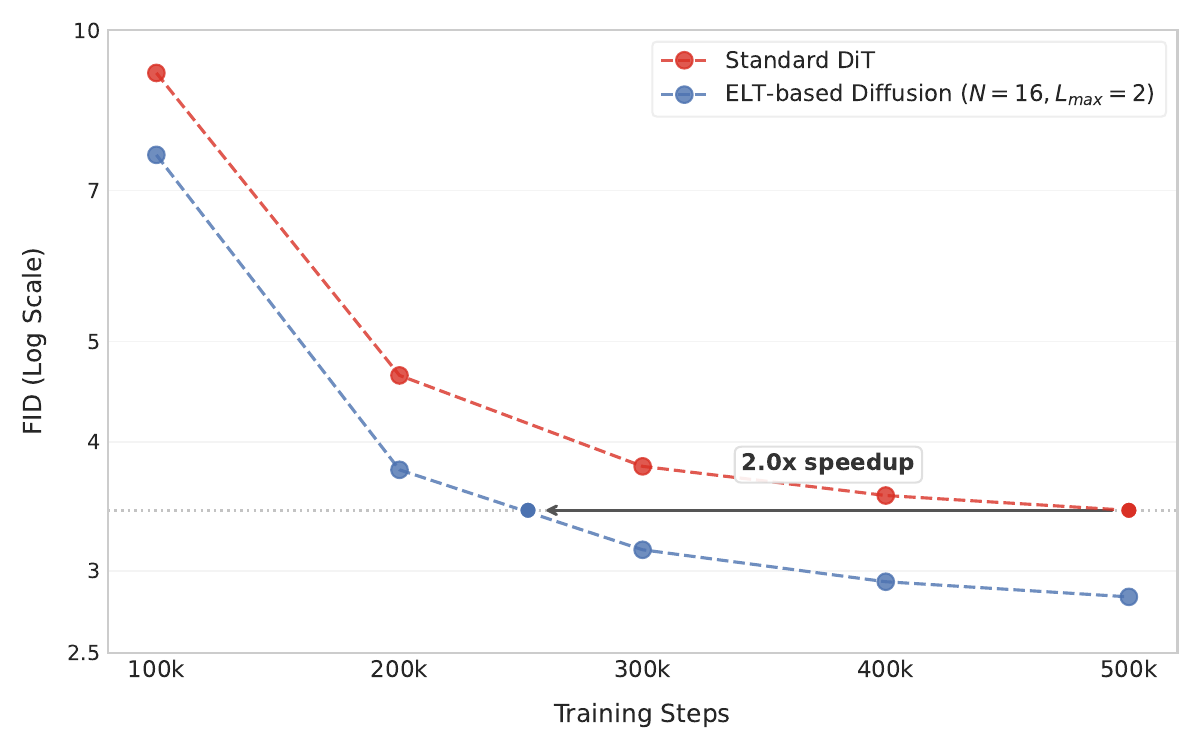}
        \label{fig:supp_16_2_faster_training_convergence}
    \end{subfigure}%
    \hfill 
    \begin{subfigure}[t]{0.48\textwidth}
        \centering
        \includegraphics[height=45mm]{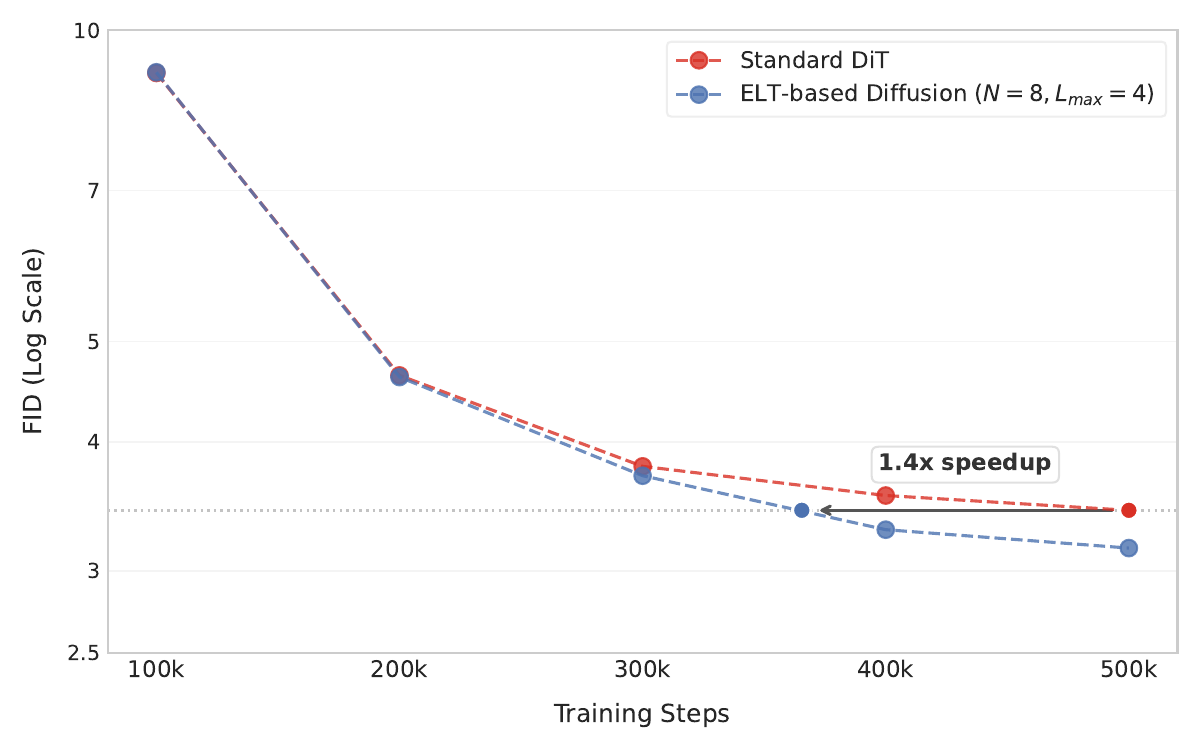}
        \label{fig:supp_8_4_faster_training_convergence}
    \end{subfigure}
    \caption{\small \textbf{Faster Training Convergence with \ours}. While using $2\times$ less parameters, \ours\ improves the training efficiency in diffusion framework.}
    \label{fig:supp_faster_training_convergence}
    \vspace{-1cm}
\end{figure}

\textbf{\flushleft{Faster Training Convergence}}: Our method demonstrates significantly faster convergence than standard DiT architectures \citep{peebles2023scalablediffusionmodelstransformers}. As illustrated in \Cref{fig:supp_faster_training_convergence}, \ours-based diffusion models achieve $2\times$ and $1.4\times$ speedups when using configurations of $16N \times 2L_{\text{max}}$ and $8N \times 4L_{\text{max}}$, respectively, in iso-inference-compute settings with $N=32$ DiT baseline. Note that effective depth $\mathcal{D}$ is same for both \ours\ and DiT baseline (32).

{\flushleft \textbf{\ours\ has high Throughput}:} \ours\ utilizes a compact set of shared parameters and maintain its major weight footprint closer to the accelerator computation unit. This avoids the cost of repeated memory transfers typically required in standard models. We evaluate the efficiency of our proposed architecture by measuring the throughput ratio relative to the baseline across various model scales with an inference batch size of 8, on 1 TPU slice. We choose the best performing \ours\ inference configuration from each model scale (refer \Cref{fig:fid_vs_params_imagenet}) and report their throughput gains in~\Cref{tab:thru}. Our method delivers significant performance gains across model scales L, XL and H. Note that these gains are contingent on the model size: as long as the shared parameters remain within memory capacity, the architecture eliminates redundant memory movement across iterations. For model scale B, the baseline (MaskGIT~\citep{chang2022maskgit}) is small enough to fit entirely within device memory capacity, eliminating the transfer penalty. Generally speaking, ELT offers extreme parameter efficiency, which would reduce significant memory transfers even for bigger models which need to be sharded along multiple devices.

\subsection{Video Generation}

\begin{figure}[t!]
    \centering
    
    \begin{minipage}[c]{0.48\textwidth}
        \centering
    \resizebox{\textwidth}{!}
    {
        \begin{tabular}{lcccccc}
            \toprule
            Method & Class & FVD $\downarrow$ & IS$\uparrow$  &  \# params & \# steps & \# GFlops\\
            \midrule
            RaMViD$^{\Delta*}$ \citep{hoppe2022diffusion}  &  & - & 21.71 $\pm$  \textcolor{gray}{0.21} & 308M & 500 & -\\
            StyleGAN-V$^{\Delta*}$\citep{skorokhodov2022stylegan}  &  & - & 23.94 $\pm$  \textcolor{gray}{0.73} & - & 1 & -\\
            DIGAN$^{\Delta}$ \citep{yu2022generating} &  & 577$\pm$\textcolor{gray}{21} & 32.70$\pm$ \textcolor{gray}{0.35} & - & 1 & $\sim$148\\
            DVD-GAN$^{\Delta}$ \citep{clark2019adversarial}  & $\checkmark$ & - & 32.97$\pm$ \textcolor{gray}{1.70} & - & 1 & -\\
            Video Diffusion$^{\Delta*}$ \citep{ho2022video}  &  &  & 57.00$\pm$ \textcolor{gray}{0.62} & 1.1B & 256 & -\\
            TATS$^{\Delta}$ \citep{ge2022longvideogenerationtimeagnostic} &  & 420$\pm$ \textcolor{gray}{18} & 57.63$\pm$ \textcolor{gray}{0.24} & 321M & 1024 & -\\
            CCVS+StyleGAN$^{\Delta}$ \citep{le2021ccvs} &  & 386$\pm$ \textcolor{gray}{15} & 24.47$\pm$ \textcolor{gray}{0.13}& - & - & - \\
            Make-A-Video$^{\Delta*}$ \citep{singer2022make}  &  $\checkmark$& 367 & 33.00 & - & - & - \\
            TATS$^{\Delta}$ \citep{ge2022longvideogenerationtimeagnostic} & $\checkmark$  & 332$\pm$ \textcolor{gray}{18} & 79.28$\pm$ \textcolor{gray}{0.38} & 321M & 1024 & - \\
            \midrule
            \textcolor{gray}{CogVideo}$^{\Delta*}$ \citep{hong2022cogvideo} & $\checkmark$ & 626 & 50.46 & 9.4B & - & - \\
            \textcolor{gray}{Make-A-Video}$^{\Delta*}$ \citep{singer2022make} & $\checkmark$  & 81 & 82.55 & $\gg$3.5B & $\gg$250 & -\\
            PAR-4$\times^{\Delta}$ \cite{wang2024parallelizedautoregressivevisualgeneration} &$\checkmark$ & 99.5 & - & 792M & 323 & -\\
            PAR-16$\times^{\Delta}$ \cite{wang2024parallelizedautoregressivevisualgeneration} &$\checkmark$ & 103.4 & - & 792M & 95 & -\\
            MaGNeTS$^{\Delta}$ \citep{goyal2025maskedgenerativenestedtransformers} & $\checkmark$ &96.4$\pm$\textcolor{gray}{2} &  88.53$\pm$\textcolor{gray}{0.20} & 306M & 12 & $\sim$1.7k \\
            \midrule
            MAGVIT-L$^{\Delta}$ \citep{yu2023magvit} & $\checkmark$ & 76$\pm$ \textcolor{gray}{2} & 89.27$\pm$ \textcolor{gray}{0.15}  & 306M & 12 & $\sim$4.3k \\ 
            
            \textbf{\ours}\ ($\bm{6N \times 4L}$) & $\checkmark$ &72.8$\pm$\textcolor{gray}{2.5} &  88.27$\pm$\textcolor{gray}{0.33} & \textbf{76M} & 12 & $\sim$4.3k \\
            
            \textbf{\ours}\ ($\bm{6N \times 6L}$) & $\checkmark$ &60.8$\pm$\textcolor{gray}{2.7} &  87.88$\pm$\textcolor{gray}{0.39} & \textbf{76M} & 24 & $\sim$13k \\
            \bottomrule
        \end{tabular}
     }
        \vspace{2mm}
        \captionof{table}{\textbf{Video Generation on UCF-101.} Methods in \textcolor{gray}{gray} are pretrained on additional large video data. $\checkmark$ denotes class-conditional. $^{*}$ indicates custom resolutions. $^\Delta$ denotes values from prior publications. No guidance is used.}
        \label{tab:ucf101_quant}
    \end{minipage}
    \hfill
    \begin{minipage}[c]{0.48\textwidth}
        \centering
        \vspace{0pt} 
        \includegraphics[width=1.0\textwidth]{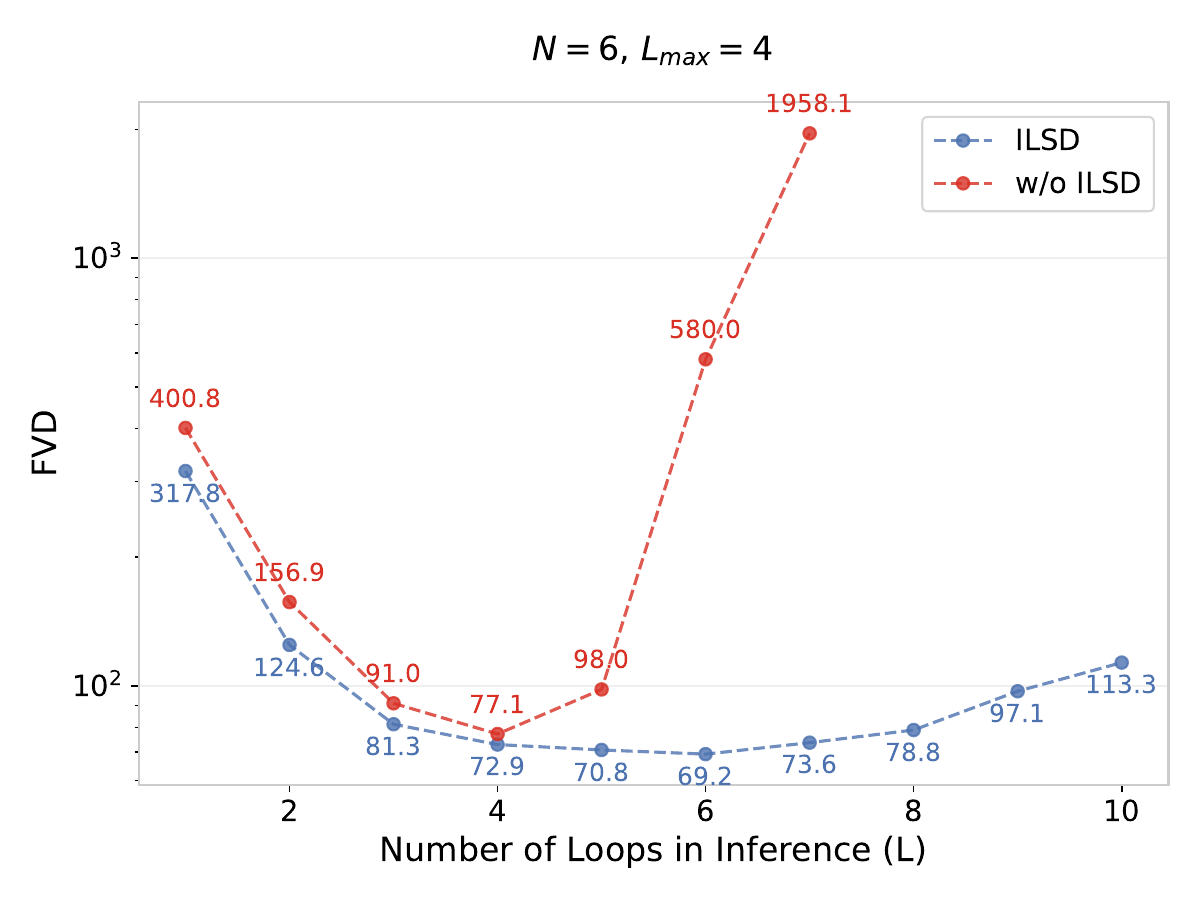}
        \captionof{figure}{\textbf{Impact of \distill\ on class-conditional UCF-101 generation.} \distill\ significantly improves performance for all values of $L$ in inference especially when $L \neq L_{max}$.} 
        \label{fig:ucf101_sjo}
    \end{minipage}
\end{figure}

We use the MAGVIT~\citep{yu2023magvit} framework to train parallel decoding based video generation models.
We summarize the results for class-conditional video generation on UCF-101 in~\Cref{tab:ucf101_quant}. 
Our compact 76M \ours\ model outperforms MAGVIT baseline (FVD 72.8 vs 76) on data-constrained settings of UCF-101 ($\sim$13.7M training tokens) in iso-inference-compute settings. Scaling the compute further with number of loops and sampling steps gives a boost in performance, reaching FVD  of 60.8. This suggests that looped transformers can exhibit robustness against overfitting in data-constrained regimes like UCF-101, effectively regularizing the learning process while maintaining the expressive capacity for high-quality generation.

\subsection{\distillfull\ drives Elasticity}
We analyze the impact of \distill\ for image generation across masked generative models (refer to~\Cref{fig:ilsd_effect_maskgit}) and diffusion models (refer to~\Cref{fig:ilsd_dit}). Models trained without \distill\ exhibit significant divergence from their fixed training depth ($L_{max}$). In contrast, \ours\ maintains stable, high-quality generation across the entire inference loop spectrum (see~\Cref{fig:gen_dit_ilsd}). We further analyse class-conditional video generation on UCF-101 in~\Cref{fig:ucf101_sjo}. Interestingly, \distill\ enables the model to maintain reasonable quality even at unseen depths ($L > L_{\text{max}}$). On UCF-101, the model achieves a peak FVD of \textit{69.20} at $L=6$ despite being trained with $L_{\text{max}}=4$, suggesting that \distill\ regularizes the iterative process sufficiently for modest extrapolation beyond training depth. We note that this extrapolation behavior warrants further investigation across datasets and scales.

\section{Conclusion}

We proposed a novel parameter-efficient approach to visual generation using recurrent transformers called \oursfull\ (\ours). Our approach achieves a strong empirical performance, similar to baselines, with $4\times$ fewer parameters in iso-inference-compute setting in both image and video generation tasks. Beyond significantly improved performance per parameter, we identified fundamental scaling properties of looped transformers: while increasing model width remains a primary driver of quality, recursive looping provides a unique ``test-time'' compute lever. Through our proposed \distillfull\ strategy, we train a single model that is performant across a variable number of iterations. This strategy effectively yields a continuous family of models from a single training run, enabling Any-Time inference where practitioners can traverse the pareto front to balance image quality and GFLOPs dynamically. 

Looking forward, we note that ELTs can potentially unlock more efficient inference for diffusion models. While existing approaches rely on the same network (and hence allocate same compute) across denoising steps, ELT can dynamically allocate compute across denoising steps, spending more compute where it matters the most. Additionally, in the context of recent one-step generative modeling paradigms such as consistency models~\citep{song2023consistencymodels} and drifting models~\citep{deng2026generative}, ELT can enable true elasticity: since there is only one sampling step, one can dynamically control the quality of the model at inference by varying the number of loops, without having to pre-determine the number of sampling steps as is the case with traditional multi-step diffusion models. We believe this paradigm of flexible, weight-efficient scaling offers a promising direction for deploying high-fidelity generative models on resource-constrained hardware. 


\clearpage

\appendix 
\newpage

\begin{center}
    \LARGE{\textbf{Appendix}}
\end{center}
\vspace{1cm} 

\begin{table}[ht!]
\centering
\small
\caption{\textbf{Implementation Details.} Hyperparameters and Configurations for Masked Generative Models (on ImageNet and UCF-101) and Diffusion Models (on ImageNet).}
\label{tab:unified_configs}
\vspace{2mm}
\begin{tabular}{lcccc}
\toprule
 & \multicolumn{2}{c}{Masked Generative Models} & & Diffusion Models \\
 \cmidrule(lr){2-3} \cmidrule(lr){5-5}
\textbf{Config} & ImageNet & UCF-101 & & ImageNet \\
\midrule
\textit{Model Input} & & & & \\
Latent shape & (1, 16, 16) & (4, 16, 16) & & (4, 32, 32) \\
Codebook Size & 1024 & 1024 & & --- \\
Embed / Latent dim & 1024 & 1024 & & 4096 \\
\midrule
\textit{Training / Optimization} & & & & \\
Batch Size & 512 & 256 & & 512 \\
Learning rate & $10^{-4}$ & $10^{-4}$ & & $10^{-4}$ \\
Optimizer & AdamW & AdamW & & AdamW \\
($\beta_1, \beta_2$) & (0.9, 0.96) & (0.9, 0.96) & & (0.9, 0.99) \\
Weight decay & 4.5e-2 & 4.5e-2 & & 0.01 \\
Warmup steps & 15K & 15K & & 10K \\
Training Epochs & 270 & 2000 & & 200 \\
Label drop (\%) & 10 & 0 & & 10 \\
Loss Function & Xent & Xent & & SigmoidELBO \\
Sampler / Prediction & --- & --- & & DDPM / $v$-pred \\

\midrule
\textit{Sampling} & & & & \\
Masking/Denoising Schedule & Cosine & Cosine & & Cosine \\
Sampling steps & 24 & 12 \& 24 & & 512 \\
CFG scale & $\checkmark$ & $\times$ & & 3.0 \\
Sampling Temp. & (0.5, 0.8) & (0.5, 0.8) & & --- \\

\bottomrule
\end{tabular}
\end{table}
\begin{table}[htp!]
\centering
\small
\caption{\textbf{Transformer Model Scaling}. Architecture details for various model sizes.}
\label{tab:model_scaling}
\vspace{2mm}
\setlength{\tabcolsep}{12pt}
\begin{tabular}{lcccc}
\toprule
Model Scale & d\_model & MLP dim & Heads &   Layers (in Baseline) \\
\midrule
S  & 384  & 1536  & 6  & 12 \\
B  & 768  & 3072  & 12 & 12 \\
L  & 1024 & 4096  & 16 & 24 \\
XL & 1152 & 4608  & 16 & 28 \\
H  & 1280 & 5120  & 16 & 32 \\
g  & 1408 & 6144  & 16 & 40 \\
G  & 1664 & 8192  & 16 & 48 \\
\bottomrule
\end{tabular}
\end{table}

\begin{table}[ht!]
\centering
\small
\caption{\small \textbf{Architecture Details} for \ours\ in Diffusion framework.}
\label{tab:model_scaling_diff}
\vspace{2mm}
\setlength{\tabcolsep}{12pt}
\begin{tabular}{lcccc}
\toprule
Model Scale & d\_model & MLP dim & Heads &   Layers (in Baseline) \\
\midrule
Small & 2048 & 8192 & 16 & 16 \\
Large & 2048 & 8192 & 16 & 32 \\
\bottomrule
\end{tabular}
\end{table}

\section{Implementation Details}

We explain our training and inference algorithms in \Cref{alg:elt_train} and \Cref{alg:elt_infer} respectively. We detail the \ours's implementation details below:
\\ \\
(a) \textbf{Masked Generative Models}: Refer \Cref{tab:unified_configs} for details of configurations and hyperparameters of \ours-based masked generative models. We experiment with several model scales (refer \Cref{tab:model_scaling} for architecture details). Note that for model scale $\geq$ H, we increase the weight decay for AdamW~\citep{loshchilov2019decoupledweightdecayregularization} optimizer by a factor of 5. We sweep over multiple classifier-free guidance (cfg)~\citep{ho2022classifierfreediffusionguidance} scales for \ours\ as well as baseline. We do not use cfg for UCF-101 class-conditional video generation. Sampling Temperature (STemp) mentioned in \Cref{tab:unified_configs} controls the randomness of the sampling from the categorical distribution
of logits. Tokens are sampled from logits/STemp. STemp is calculated as:
\begin{equation}
    \text{STemp} = bias + scale \times (1 - (k+1)/K)
\end{equation}
where $bias$ and $scale$ are hyperparameters, $k$ is current sampling step and $K$ is total number of sampling steps. Throughout our experiments, we use $bias = 0.5$ and $scale = 0.8$.
\\ \\
(b) \textbf{Diffusion Models}:  Refer \Cref{tab:unified_configs} for details of configurations and hyperparameters of \ours-based diffusion models. For diffusion models, we experiment with two model architectures (refer \Cref{tab:model_scaling_diff} for details).

\begin{algorithm}[htbp!]
\caption{\textbf{\ours\ Training with \distillfull}}
\label{alg:elt_train}
\vspace{1mm}
\hrule
\vspace{1mm}
\begin{lstlisting}[mathescape=true]
# $g_{\Theta}$: weight-shared composite transformer block
# $head$: task-specific prediction head
# $x$, $y$: input features and ground truth target
# $L_{max}$: maximum loop count for training
# $L_{min}$: minimum loop count for student sampling
# $\lambda$: curriculum weight (decays 1 $\rightarrow$ 0 over training)

# 1. Sample intermediate student loop count
$L_{int}$ = jax.random.randint(key, shape=(), minval=$L_{min}$, maxval=$L_{max}$)

# 2. Define the shared forward step
def loop_body(step, states):
    $F_{curr}$, $F_{int}$ = states
    
    # Apply the shared block exactly once per step
    $F_{curr}$ = $g_{\Theta}$($F_{curr}$)
    
    # Save the intermediate state exactly when we reach $L_{int}$ loops (0-indexed).
    $F_{int}$ = jax.lax.cond(
        step == $L_{int}$ - 1,
        lambda _: $F_{curr}$,  # Store the intermediate feature
        lambda _: $F_{int}$,   # Retain the previously initialized feature
        None
    )
    return $F_{curr}$, $F_{int}$

# 3. Execute forward pass
# Initialize both tracking variables with $x$
$F_{max}$, $F_{int}$ = jax.lax.fori_loop(0, $L_{max}$, loop_body, ($x$, $x$))

# 4. Compute Task Losses (Ground-truth)
$\mathcal{L}^{GT}_{max}$ = $\mathcal{L}^{GT}$($head$($F_{max}$), $y$)
$\mathcal{L}^{GT}_{int}$ = $\lambda$ * $\mathcal{L}^{GT}$($head$($F_{int}$), $y$)

# 5. Compute ILSD Distillation Loss
$\mathcal{L}^{distill}$ = (1 - $\lambda$) * $\mathcal{L}^{dist}$($F_{int}$, jax.lax.stop_gradient($F_{max}$))

# 6. Total Loss
loss = $\mathcal{L}^{GT}_{max}$ + $\mathcal{L}^{GT}_{int}$ + $\mathcal{L}^{distill}$
\end{lstlisting}
\vspace{-1mm}
\hrule
\end{algorithm}
\begin{algorithm}[htbp]
\caption{\textbf{\ours's Any-Time Inference}}
\label{alg:elt_infer}
\small{\textbf{Note:} Algorithm is for a single sampling step. Compute is dynamically scalable by adjusting the loop budget $L$, allowing for flexible inference-time compute without model retraining.}
\vspace{1mm}
\hrule
\vspace{1mm}
\begin{lstlisting}[mathescape=true]
# $g_{\Theta}$: weight-shared composite transformer block
# $head$: task-specific prediction head 
# $x$: input features for the current generation step
# $L$: dynamic compute budget (number of loops to execute)

$F$ = $x$

# Recursive refinement using weight-shared parameters
for step in range($L$):
    $F$ = $g_{\Theta}$($F$)

# Single exit through the prediction head
out = head($F$)

return out
\end{lstlisting}
\vspace{-1mm}
\hrule
\end{algorithm}

\section{Additional Experiments}
{\flushleft \textbf{Cost and Efficiency}:} ELT drastically reduces the unique param count, which in turn helps in following: (i) \textbf{\textit{Hardware Costs}}: The primary bottleneck in training and deploying modern generative models is often memory (HBM/VRAM). ELT reduces the memory budget needed to train and serve larger effective depths. (ii) \textbf{\textit{Training Speedup}}: As ELT trains intermediate loops using a single forward pass with negligible cost of the prediction head, training step time remains same. 
\Cref{tab:train_analyse} (corresponding to \Cref{tab:dit} results) profiles DiTs training on TPU v6e with 8N × 4L looping config, using a TPU v6e with $8N\times4L$ looping config.
$\mathcal{D}$ (effective depth) is 32 for all methods. This infrastructure efficiency compounds with the algorithmic convergence speed-up demonstrated in \Cref{fig:supp_faster_training_convergence}, these complementary benefits yield a overall reduction in training wall-clock time.
\begin{table}[htbp]
    \centering
    \caption{\small \textbf{Training Profile}.}
    \setlength{\tabcolsep}{2pt}
\begin{tabular}{lcc}
    \toprule
    Method & \makecell{Step \\ Time (ms)} & \makecell{Peak \\ HBM (gb)} \\
    \midrule
    Baseline (32 layers) & 130 & 7.7 \\
    Just Looping & 130 & \textbf{5.1} \\
    ELT (w/ ILSD)   & 130 & \textbf{5.1} \\
    ELT$^{opt}$ (w/ ILSD) & \textbf{118} & 5.8 \\
    \bottomrule
\end{tabular}
\label{tab:train_analyse}
\vspace{-10pt}
\end{table}
 While matching the baseline's theoretical FLOPs, ELT inherently reduces peak HBM memory by 34\%, which acts as a \textbf{budget to speedup training}. For instance, ELT$^{opt}$ uses unrolling (``for loop") instead of ``jax.lax.scan" across layers axis. This generates a new computational graph for every layer, removing some sequential computation, utilizing the ELT's memory savings and reducing step time. Other orthogonal optimizations include increasing batch-size and gradient checkpointing. (iii) \textit{\textbf{Inference Speedup}}: Please refer to \Cref{tab:thru} for quantitative comparison. ELT eliminates redundant memory transfers (HBM $\rightarrow$ VRAM), achieving \textbf{3.5x} peak throughput speedup in \textit{iso-inference compute} settings.

\begin{figure}[h!]
\vspace{-8pt}
    \centering
    \caption{\small \textbf{Losses Ablation}. LHS: DiTs, RHS: Masked Generative.}
    \begin{minipage}{0.48\textwidth}
        \centering
        \includegraphics[width=\linewidth]{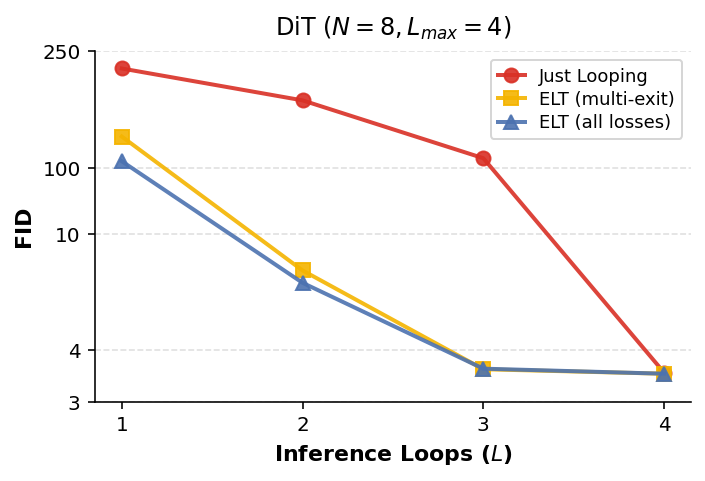}
    \end{minipage}\hfill
    \begin{minipage}{0.48\textwidth}
        \centering
        \includegraphics[width=\linewidth]{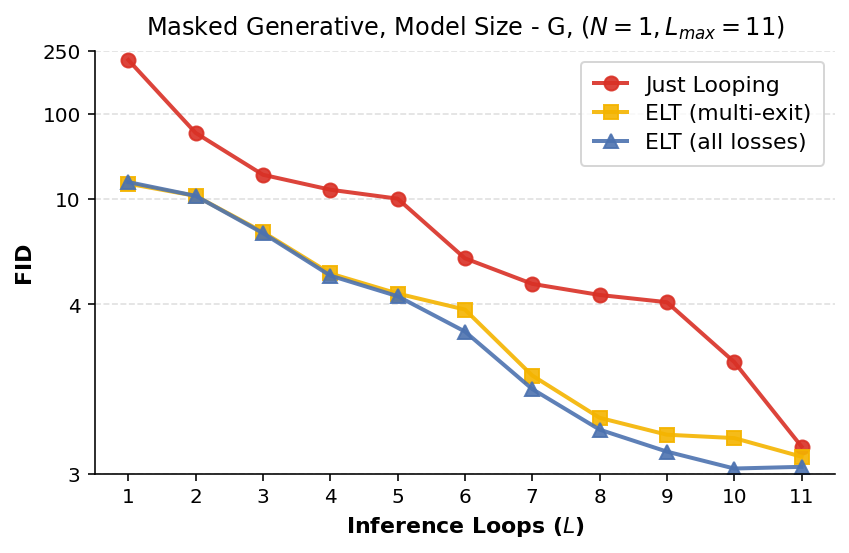} 
    \end{minipage}
    \label{fig:loss_ablate}
\end{figure}

 \textbf{\flushleft{Losses Ablation}}: \Cref{fig:ucf101_sjo} (video gen) and \Cref{fig:distill_impact_combined} (image gen) show the ablation of vanilla looping vs ELT across masked generative models and DiTs. We further ablate the importance of ILSD loss in \Cref{fig:loss_ablate}.

\begin{figure}
    \centering
    \caption{\small \textbf{Iterative Refinement.}}
    \includegraphics[width=\textwidth]{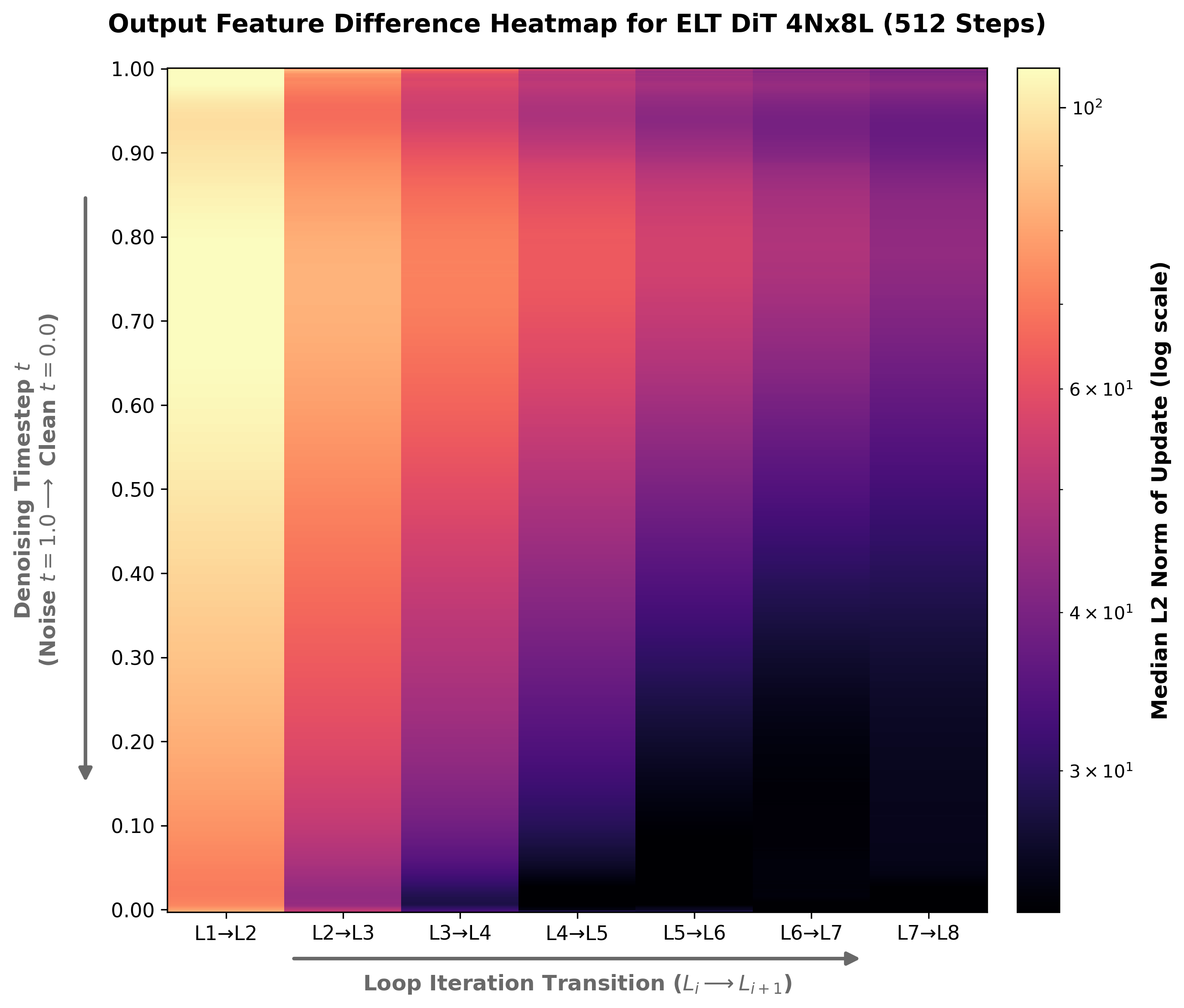}
    \label{fig:inter}
\end{figure}

\textbf{\flushleft{Iterative Refinement}}: We analyzed the per-iteration update magnitudes (L2 norm) across the loops and sampling steps (\Cref{fig:inter}). It shows that initial denoising steps have more updates for higher loops than later denoising steps. These observations can potentially be used to make the number of loops adaptive across denoising steps. 

\section{Limitations}

While \ours\ demonstrates strong parameter efficiency and elastic inference capabilities, we identify some limitations of the current work. 

{\flushleft \textbf{Failure Cases}:} We observe that \ours\ performance degrades in two notable scenarios: (i) when the number of unique layers $N$ is too small (e.g., $1N \times 32L$ configuration in \Cref{tab:dit}), a single layer lacks representational capacity regardless of loop count, and (ii) when loop count $L$ significantly exceeds $L_{\text{max}}$ at inference, quality can deteriorate as the shared block over-iterates beyond its trained convergence regime.

{\flushleft \textbf{Deployment Considerations}:} While \ours\ offers significant parameter savings and throughput gains, practical deployment requires careful selection of the operating point ($N$, $L$) based on the target hardware. The optimal loop count $L$ for a given quality target is model-scale dependent and should be calibrated per deployment tier.

\section{Acknowledgments}
We thank Nikunj Saunshi, Thomas Mensink and Ashwini Pokle for their constructive comments and valuable suggestions, which enhanced the overall quality of this manuscript.

\end{document}